\documentclass[11pt]{article}

\usepackage{acl}

\usepackage[T1]{fontenc}
\usepackage[utf8]{inputenc}

\usepackage{xcolor}
\usepackage{hyperref}
\hypersetup{
    colorlinks=true,
    linkcolor=blue,
    filecolor=magenta,
    urlcolor=cyan,
}

\usepackage{times}
\usepackage{latexsym}

\usepackage{graphicx}
\graphicspath{{latex/}{./}{../}}
\usepackage{booktabs}
\usepackage{multirow}
\usepackage{adjustbox}
\usepackage{makecell}
\usepackage{tabu}
\usepackage{color, colortbl}
\usepackage{enumitem}
\usepackage{placeins}
\usepackage{float}

\usepackage{amsmath}
\usepackage{amssymb}
\usepackage{algorithm}
\usepackage{algpseudocode}

\usepackage{tikz}
\usetikzlibrary{arrows.meta,positioning,fit,calc}
\usetikzlibrary{backgrounds}

\usepackage{array}
\newcolumntype{C}[1]{>{\centering\arraybackslash}p{#1}}

\definecolor{systemcolor}{HTML}{E8F1FF}
\definecolor{usercolor}{HTML}{F7FFE9}
\definecolor{assistantcolor}{HTML}{FFF7E9}

\usepackage{tikz}
\usepackage{pgfplots}
\pgfplotsset{compat=1.18}
\usetikzlibrary{fit, positioning}

\usepackage[most]{tcolorbox} 
\tcbuselibrary{breakable,skins}
\usepackage{tikz}
\usepackage{xcolor}
\usepackage{enumitem}
\usepackage{booktabs}
\usepackage{multirow}
\usepackage{microtype}

\newcommand{\ours}{\textbf{HiKEY}}

\newcommand\blfootnote[1]{%
  \begingroup
  \renewcommand\thefootnote{}\footnote{#1}%
  \addtocounter{footnote}{-1}%
  \endgroup
}

\title{HiKEY: Hierarchical Multimodal Retrieval for Open-Domain Document  Question Answering}

\author{%
  Joongmin Shin$^{1}$,\ Gyuho Shim$^{3}$,\ Jeongbae Park$^{1}$,\ Jaehyung Seo$^{2\ddagger}$,\ Heuiseok Lim$^{1,3\ddagger}$ \\[0.35em]
  \normalfont
  $^{1}$Human-inspired AI Research, Korea University \\
  $^{2}$Computer Science and Engineering, Konkuk University \\
  $^{3}$Department of Computer Science and Engineering, Korea University \\[0.35em]
  \texttt{\{tlswndals13, gjshim, insmile, limhseok\}@korea.ac.kr} \\
  \texttt{seojae777@konkuk.ac.kr}
}

\begin{document}
\maketitle
\blfootnote{$^{\ddagger}$Corresponding author.}

\begin{abstract}
Retrieval-augmented generation (RAG) for document-based Open-domain Question Answering (ODQA) on large-scale industrial corpora faces two critical bottlenecks: routing failure in locating the correct document and evidence fragmentation in integrating scattered information.
Existing approaches relying on flat text chunks or page-level images inherently struggle to (i) precisely pinpoint the target document among thousands of candidates and (ii) organically connect multimodal evidence, such as tables and figures, within a limited token budget.
To address these challenges, we propose \ours{}, a hierarchical tree-based multimodal retrieval framework that elevates document hierarchy to a first-class retrieval signal.
Instead of simple chunking, \ours{} reconstructs a logical heterogeneous graph via Document Hierarchical Parsing (DHP), explicitly encoding parent--child relationships.
Adopting a hierarchical coarse-to-fine strategy, the framework (1) performs global routing to rapidly prune the search space using hierarchical indexing, and (2) conducts fine-grained retrieval to rank sections by employing a multimodal fusion strategy that captures the most discriminative evidence.
Finally, \ours{} assembles a token-efficient evidence subgraph via a hybrid structural-semantic packing strategy.
Experiments on ODQA benchmarks demonstrate that \ours{} significantly outperforms page- and chunk-based baselines, improving retrieval recall by up to 12.9\% and end-to-end QA performance by up to 6.8\%.
\end{abstract}

\section{Introduction}
\label{sec:intro}

Retrieval-augmented generation (RAG) has become an essential technique for handling large, information-dense document corpora in expert domains such as finance, law, and engineering~\cite{lewis2021retrievalaugmented, Jeong2023ASO, Ge2023Development}.
In real industrial settings, documents are rarely simple text streams; rather, they typically span dozens of pages (multi-page) and contain a complex mixture of multimodal elements such as tables, figures, schematics, captions, and footnotes~\cite{Gong2020Recurrent, qu-etal-2025-semantic, tito2023hierarchicalmultimodaltransformersmultipage}.
Beyond within-document retrieval, Open-domain Question Answering (ODQA) requires identifying the correct documents and sections from a large corpus. In such settings, system-level performance and cost are determined not only by the generator, but also by the choice of retrieval unit and the efficiency of the exploration procedure (routing and aggregation)~\citep{gao2024retrievalaugmentedgenerationlargelanguage}.

The most widely adopted approach is text chunk-based RAG, which splits documents into text chunks and retrieves the top-$k$ chunks by similarity~\cite{Gong2020Recurrent, duarte-etal-2024-lumberchunker}.
However, this approach fails to reflect key properties of complex industrial documents.
First, the evidence for an answer is often not self-contained within a single paragraph, but fragmented across multiple pages or documents.
Second, evidence is entangled with section hierarchies (e.g., 1 $\rightarrow$ 1.2 $\rightarrow$ 1.2.1) and explicit in-text pointers such as ``Table 2'' or ``Figure 3''.
Third, crucial information is frequently embedded in non-text regions such as tables and charts~\cite{DocHieNet}.
Most importantly, text chunks are confined to local context and can easily miss global structural signals such as the table of contents or section paths.
This often results in (1) routing failure, where the system cannot find the correct document or section during corpus exploration, or (2) incomplete answers due to disconnected evidence~\cite{hong-etal-2024-intelligent, multidocfusion}.

To overcome these limitations, GraphRAG methods that leverage connectivity between chunks have been proposed~\cite{sarthi2024raptorrecursiveabstractiveprocessing,liu-etal-2025-hoprag}, but most of them focus on text based links and fail to fully incorporate multimodal elements.
As an alternative, multimodal RAG that embeds full-page images has gained attention~\cite{faysse2025colpali, Cho_2025_ICCV, vdocrag2025}, but Page-level units inject large amounts of irrelevant content, dramatically increasing token costs, and make it difficult to fit information from many documents within a limited context window~\citep{ma-etal-2025-towards-storage}.

We argue that the core to industrial ODQA is not merely expanding modality, but redesigning the retrieval unit and exploration procedure using document hierarchy.
Accordingly, motivated by the intuition that hierarchical structure is the key to multimodal ODQA, we propose \ours{}, which exploits structural information as a first-class retrieval signal.
Specifically, \ours{} (1) reconstructs a hierarchical tree from unstructured documents and redefines paragraphs, tables, and figures as multimodal evidence units annotated with section paths, and (2) builds a hierarchical graph that explicitly preserves the logical document structure.

During retrieval, \ours{} follows a hierarchical coarse-to-fine strategy, analogous to how humans read long documents:
it first uses hierarchical cues to quickly route to a small set of candidate documents, then ranks sections within the routed documents by scoring their multimodal units (text/table/figure) using a combination of lexical, dense text-semantic, and dense visual-semantic signals to capture the most discriminative evidence within the hierarchical context.
Finally, \ours{} assembles an evidence subgraph under a fixed token budget by employing a hybrid packing strategy that prioritizes sibling nodes and semantic associates, enabling coherent integration of scattered evidence.

\paragraph{Contributions}
\begin{itemize}[leftmargin=*]
    \item \ours{}: We propose a hierarchical tree-based multimodal retrieval framework that elevates document hierarchy from passive metadata to a first-class retrieval signal. By reconstructing a logical graph via DHP, \ours{} structurally resolves the evidence fragmentation problem inherent in industrial documents, bridging the gap between flat retrieval and complex document layouts.
    
    \item \textbf{Hierarchical Coarse-to-Fine Retrieval}: We introduce a novel retrieval strategy that integrates global routing with local fine-grained ranking. Specifically, we employ a fine-grained multimodal fusion strategy that identifies the most discriminative evidence within a section, coupled with a graph traversal mechanism that enriches visual units with their hierarchical upper context. This ensures robustness against visual ambiguity and missing captions.
    
    \item \textbf{Comprehensive Experiments}: Across industrial-aligned multi-page document ODQA benchmarks, \ours{} consistently outperforms state-of-the-art text-based and multimodal RAG baselines. Our method improves retrieval performance by 4.5\%--12.9\% and QA performance by 3.7\%--6.8\%, validating the efficacy of our ancestry-aware subgraph assembly in token-limited settings.
\end{itemize}

\begin{figure*}[ht!]
\begin{center}
\includegraphics[width=1.0\linewidth]{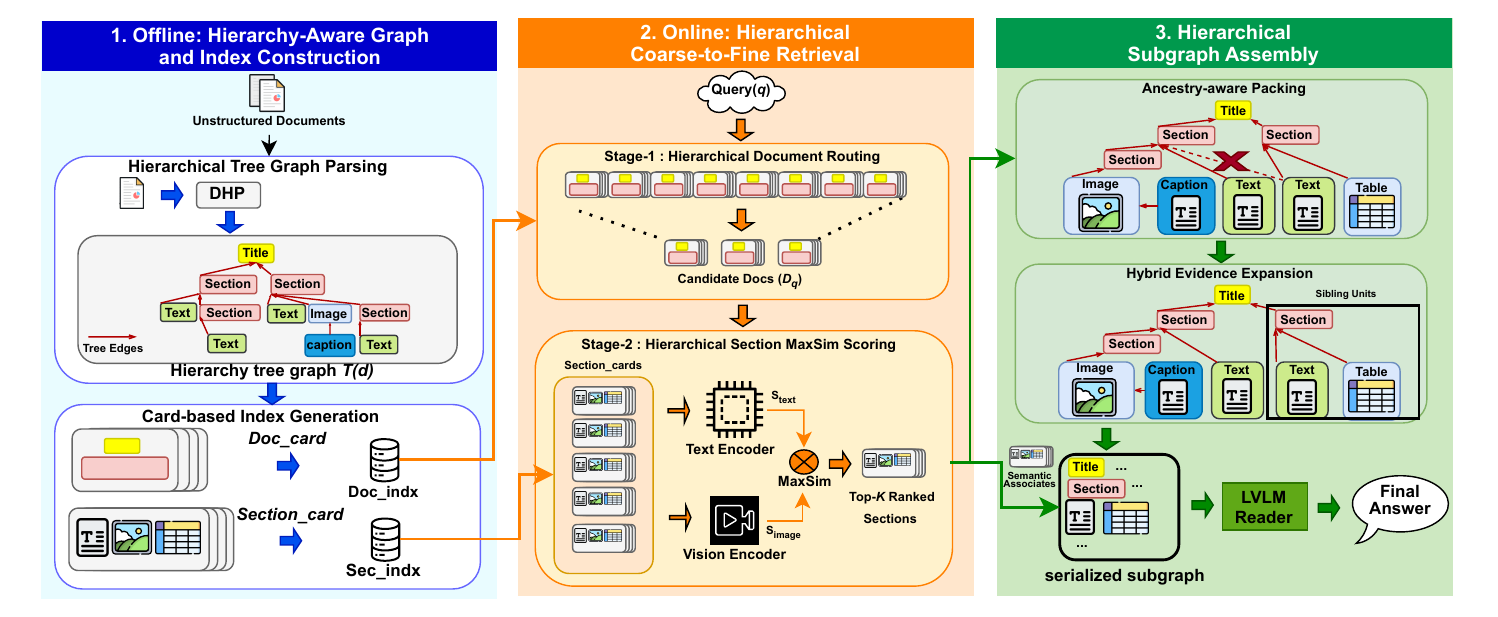}
\end{center}
\caption{Overview of \ours{}, showing 1.~offline hierarchy-aware graph and index construction; 2.~online hierarchical coarse-to-fine retrieval (Stage-1 document routing followed by Stage-2 section scoring); 3.~ancestry-aware evidence subgraph assembly for an LVLM reader.}
\label{fig:system_architecture}
\end{figure*}

\section{Related Work}

\paragraph{Chunking strategies and retrieval units for RAG.}
RAG performance depends on the retrieval unit and chunking strategy~\citep{gao2024retrievalaugmentedgenerationlargelanguage}, ranging from fixed-length segmentation~\citep{Gong2020Recurrent} to semantic or structure-aware chunking~\citep{duarte-etal-2024-lumberchunker, zhao-etal-2025-moc, yepes2024financialreportchunkingeffective,verma2025s2chunkinghybridframework}.
However, text-centric units often miss non-text evidence (tables/figures) and provide weak global structural cues for corpus-level routing in ODQA, even with hierarchical chunking~\citep{multidocfusion}.

\paragraph{Page-level multimodal retrieval.}
To compensate for the limitations of text chunking, visual approaches that embed full-page images have been introduced~\citep{faysse2025colpali, Cho_2025_ICCV, vdocrag2025}.
While they preserve layout information, they suffer from a structural drawback: coarse granularity.
First, pages contain substantial irrelevant content that acts as noise and reduces retrieval precision.
Second, using full-page images as input in multi-document settings drastically increases token costs and becomes inefficient~\citep{ma-etal-2025-towards-storage}.
Moreover, simply concatenating pages makes it difficult for models to capture a document's logical flow (section hierarchy) and the explicit links between distant evidence.

\paragraph{Graph-based retrieval.}
GraphRAG connects information via graphs to support deeper contextual reasoning beyond simple similarity search.
Prior work extracts entity relations to build knowledge graphs~\citep{he2024gretrieverretrievalaugmentedgenerationtextual} or summarizes and links text chunks for hierarchical exploration~\citep{sarthi2024raptorrecursiveabstractiveprocessing,liu-etal-2025-hoprag}.
However, these methods mainly rely on text nodes and cannot use multimodal evidence (tables/figures) as first-class retrieval units.
Recent work proposes multimodal graphs with pages as nodes~\citep{wu-etal-2025-molorag, jain-etal-2025-simpledoc}, but page-level granularity still limits fine-grained evidence assembly and token efficiency.

Overall, existing approaches have not fully resolved the trade-off between coverage (not missing answer documents) and efficiency (assembling complete evidence under limited resources).
To address this, we propose \ours{}, which integrates (1) fine-grained multimodal units grounded in DHP-based hierarchy, (2) a hierarchical coarse-to-fine strategy that separates global routing from local ranking, and (3) an ancestry-aware subgraph assembly that dynamically connects scattered evidence via hybrid structural and semantic association.

\paragraph{Industrial document QA and search.}
Enterprise deployment of Document QA and Search (corpus-level ODQA) is characterized by large repositories of long, multi-page PDFs with mixed modalities (text, tables, figures), complex hierarchies, and limited Reader token budgets -- the same operating regime we target in this paper.
Prior production-leaning studies focus on individual pieces of this stack: long-context retrieval over enterprise-scale collections~\citep{ma-etal-2025-towards-storage, multidocfusion}, layout- and structure-aware chunking for financial/technical documents~\citep{yepes2024financialreportchunkingeffective, verma2025s2chunkinghybridframework, hong-etal-2024-intelligent}, and multimodal retrieval that preserves page layout~\citep{faysse2025colpali, Cho_2025_ICCV, vdocrag2025}.
We complement this literature by connecting corpus-level routing, block-level multimodal retrieval, and budget-aware packing into a single framework, and by making the industrial-aligned analyses that matter in deployment first-class evaluation axes: strict routing (\emph{All@K}), evidence-type and multi-hop breakdowns, and token-budget sensitivity (Tables~\ref{tab:doc_retrieval_avg_1_10},~\ref{tab:mmqa_method_breakdown_source_hop}, and \ref{tab:cmp_budget_sensitivity_with_baselines}).

\paragraph{Task scope.}
This paper studies \emph{corpus-level ODQA retrieval over large PDF collections}: \ours{} performs hierarchical document routing and section/unit retrieval to assemble evidence under a fixed Reader budget, and we report document identification (Recall/MRR/Hit/All) and end-to-end QA (EM/ANLS/ROUGE-L/METEOR). We do not generate keyphrase lists nor evaluate against keyphrase ground truth (e.g., keyphrase F1@K). While our ODQA benchmarks are open-domain (multi-topic) Wikipedia-style text and layout, our DHP hierarchy backbone is separately validated on scanned/pixel and explicitly multi-domain parsing benchmarks (Appendix~\ref{sec:dhp_details}, Tab.~\ref{tab:dhp_backbone_validation_appendix}), supporting robustness across document formats and domains.

\section{\ours{}}
\label{sec:method}

\ours{} is designed to resolve inefficient corpus exploration and evidence fragmentation in ODQA.
It is inspired by how humans search long documents: they first skim the table of contents to narrow the scope (routing), and then carefully read the relevant sections and associated tables/figures (fine retrieval).
As shown in Fig.~\ref{fig:system_architecture}, the framework consists of two major phases: (1) Offline Hierarchy-Aware Graph and Index Construction, and (2) Online Hierarchical Coarse-to-Fine Retrieval.

\paragraph{Notation.}
Let $\mathcal{D}$ be a corpus of documents and $q$ a user query.
Each document $d \in \mathcal{D}$ is parsed into a hierarchy tree $\mathcal{T}(d)$.
We define two types of index cards: (i) a Doc\_card $x_d$ representing the global document context, and (ii) a Sec\_card $x_s$ containing a set of multimodal units $C_s = \{c_1, c_2, \dots, c_n\}$ for each section $s \in \mathcal{T}(d)$.
Each unit $c$ is assigned a type $\tau(c) \in \{\text{Text, Table, Image}\}$.

\subsection{Offline: Hierarchy-Aware Graph and Index Construction}
To address the challenge of fragmented evidence highlighted in Sec.~\ref{sec:intro}, we transform each document $d$ from a raw format into a structural heterogeneous graph and a corresponding searchable index.

\paragraph{Hierarchical Tree Graph Parsing.}
Conventional RAG often loses global context because it splits documents into small, flat chunks.
To mitigate this, we use the Document Hierarchical Parsing (DHP) module from M3DocDep~\citep{shin2026m3docdep}, which is trained on public hierarchy parsing datasets, to reconstruct a logical hierarchy tree $\mathcal{T}(d)$ from page-level layout blocks.
Each block (text, table, figure) becomes a node, connected by parent--child Tree Edges ($E_{\text{tree}}$), and receives an ancestor section path (e.g., \texttt{Title > Section 5 > 5.3}).

\paragraph{Card-based Index Generation.}
We generate hierarchical cards for retrieval:
\begin{itemize}[leftmargin=*]
    \item Doc\_card ($x_d$): title plus section paths for global routing, $x_d = \texttt{Title}(d) \Vert \bigcup_{s \in \mathcal{T}(d)} \texttt{path}(s)$.
    \item Sec\_card ($x_s$): a section-level card containing its text and non-text units (tables/figures) as searchable items.
\end{itemize}

\paragraph{Contextual Enrichment via Graph Traversal.}
For visual units $c \in \{\text{Table, Image}\}$ lacking captions, \ours{} traverses the DHP tree to retrieve a logically preceding textual node as Upper Context, stored as $\text{ctx}(c)$ to reduce visual ambiguity.

\subsection{Online: Hierarchical Coarse-to-Fine Retrieval}
Given a query $q$, \ours{} identifies the most relevant evidence by traversing the hierarchical search space from the document level down to the multimodal unit level.

\paragraph{Stage-1: Hierarchical Document Routing.}
To narrow the search scope within a vast corpus, we first identify a candidate set of documents $\hat{\mathcal{D}}_q$ by ranking Doc\_cards using a hybrid score:
{\small
$$S_{\text{doc}}(d, q) = \alpha \cdot \mathrm{mm}(s_{\text{lex}}(x_d, q)) + (1-\alpha) \cdot \mathrm{mm}(s_{\text{text}}(x_d, q))$$
}
where $\mathrm{mm}(\cdot)$ denotes per-query min--max normalization. This stage filters out irrelevant documents globally, leveraging the hierarchical paths indexed in the Doc\_cards.

\paragraph{Stage-2: Hierarchical Section MaxSim Scoring.}
Within $\hat{\mathcal{D}}_q$, we perform fine-grained retrieval to rank all constituent Sec\_cards globally. We define a Type-Specific Unit Score $s(c, q)$ to leverage modality-specialized encoders:
\begin{equation*}
\small
\resizebox{\columnwidth}{!}{$
s(c, q)=
\begin{cases}
s_{\text{hybrid}}(\text{text}(c), q) & \tau(c)=\text{Text}\\
\gamma s_{\text{vis}}(c, q) + (1-\gamma) s_{\text{hybrid}}(\text{ctx}(c), q)
& \tau(c)\in\{\text{Table},\text{Image}\}
\end{cases}
$}
\end{equation*}

The relevance of a section is determined by the MaxSim operator~\citep{10.1145/3397271.3401075} over its units: $S_{\text{sec}}(s, q) = \max_{c \in C_s} s(c, q)$. The final section score for ranking is fused as: $S_{\text{final}}(s, q) = \lambda S_{\text{doc}}(\text{doc}(s), q) + (1-\lambda) S_{\text{sec}}(s, q)$.

\subsection{Hierarchical Subgraph Assembly}
Finally, \ours{} assembles the top-$K_{\text{sec}}$ ranked sections into a serialized subgraph for the Reader model.
To maximize information density within a fixed token budget $B_{\text{tok}}$, we propose a structured assembly process that prioritizes logical coherence and evidence connectivity.

\paragraph{Ancestry-aware Packing.}
For each selected section, we insert the Anchor Unit (the unit achieving the MaxSim score) and its Governing Headers (ancestral section titles in the DHP tree) to preserve logical coherence and prevent scope misinterpretation.

\paragraph{Hybrid Evidence Expansion.}
To effectively associate scattered evidence without predicting an explicit cross-block link graph, we employ a hybrid packing strategy to fill the remaining token budget:
\begin{itemize}[leftmargin=*]
    \item Sibling Units (Structural Association): We first prioritize tables and figures sharing the same parent section as the anchor. This preserves local context and visual-to-text alignment within the same subtree.
    \item Semantic Associates (Semantic Association): If the budget permits, we further expand the subgraph with non-local visual units that exhibit high vector similarity to the anchor. This captures latent cross-section dependencies, ensuring the Reader can integrate distributed multimodal evidence.
\end{itemize}
The two expansion modes are complementary: structural association recovers evidence that is guaranteed to be co-referential by document layout (same subtree), while semantic association recovers evidence that the layout fails to link but that is topically aligned with the anchor.
An algorithm-aligned schematic of these phases (Fig.~\ref{fig:packing_schematic}), a compact running example on the Apollo~11 query, and the exact Reader serialization format are provided in Appendix~\ref{sec:packing_details}, so that the reader can see \emph{what the subgraph contains, how it is constructed, and how it is interpreted}.

\newcolumntype{C}[1]{>{\centering\arraybackslash}p{#1}}
\newcommand{\wS}{0.85cm}
\newcommand{\gapA}{5pt}

\begin{table*}[t]
\centering
\scriptsize
\renewcommand{\arraystretch}{0.95}
\setlength{\tabcolsep}{1.8pt}

\begin{adjustbox}{max width=\textwidth, keepaspectratio}
\begin{tabular}{@{}l
*{4}{C{\wS}} @{\hspace{\gapA}}
*{4}{C{\wS}} @{\hspace{\gapA}}
*{4}{C{\wS}}@{}}
\toprule
 & \multicolumn{4}{c}{M3DocVQA}
 & \multicolumn{4}{c}{FRAMES}
 & \multicolumn{4}{c}{AVG} \\
\cmidrule(lr){2-5}\cmidrule(lr){6-9}\cmidrule(lr){10-13}
Method &
Recall & MRR & Hit & All &
Recall & MRR & Hit & All &
Recall & MRR & Hit & All \\
\midrule

\multicolumn{13}{l}{\textit{Text chunk-based RAG}} \\
\quad Page & 80.6 & 91.3 & 95.0 & 66.4  & 65.4 & 91.9 & 97.1 & 34.4  & 73.0 & 91.6 & 96.1 & 50.4 \\
\quad Length chunking  & 82.0 & 90.6 & 95.0 & 69.1  & 65.3 & 92.8 & 96.9 & 33.8  & 73.7 & 91.7 & 96.0 & 51.4 \\
\quad LumberChunker    & 81.5 & 90.9 & 95.1 & 68.1  & 64.8 & 92.4 & 96.8 & 33.2  & 73.2 & 91.7 & 95.9 & 50.6 \\
\quad Meta Chunker     & 81.2 & 90.8 & 95.0 & 67.7  & 64.9 & 92.4 & 96.8 & 33.3  & 73.1 & 91.6 & 95.9 & 50.5 \\
\quad Structural chunking & 80.6 & 90.5 & 94.8 & 66.9  & 64.0 & 92.1 & 96.6 & 32.4  & 72.3 & 91.3 & 95.7 & 49.7 \\
\quad MultiDocFusion   & 83.0 & 91.5 & 95.4 & 70.5  & 66.2 & 93.0 & 97.2 & 34.8  & 74.6 & 92.2 & 96.3 & 52.6 \\

\midrule
\multicolumn{13}{l}{\textit{Text-based GraphRAG}} \\
\quad RAPTOR           & 81.8 & 90.9 & 95.0 & 68.4  & 64.9 & 92.0 & 96.6 & 33.3  & 73.4 & 91.5 & 95.8 & 50.9 \\
\quad HopRAG           & 82.3 & 91.1 & 95.2 & 69.3  & 65.8 & 92.2 & 96.8 & 34.2  & 74.1 & 91.7 & 96.0 & 51.8 \\

\midrule
\multicolumn{13}{l}{\textit{Page-level multimodal RAG}} \\
\quad M3DocRAG         & 75.6 & 81.8 & 89.9 & 61.4  & 56.8 & 83.2 & 91.9 & 25.0  & 66.2 & 82.5 & 90.9 & 43.2 \\
\quad VDocRAG          & 77.1 & 83.3 & 90.9 & 63.2  & 58.5 & 84.2 & 92.5 & 26.5  & 67.8 & 83.8 & 91.7 & 44.9 \\

\midrule
\multicolumn{13}{l}{\textit{Multimodal GraphRAG}} \\
\quad MoLoRAG          & 76.0 & 82.4 & 90.2 & 62.0  & 57.4 & 83.5 & 92.1 & 25.5  & 66.7 & 83.0 & 91.2 & 43.8 \\
\quad SimpleDoc        & 77.3 & 83.5 & 91.1 & 63.6  & 60.6 & 85.4 & 93.2 & 28.4  & 69.0 & 84.5 & 92.2 & 46.0 \\

\midrule
\quad \ours{} &
\textbf{84.9} & \textbf{91.8} & \textbf{96.1} & \textbf{73.6} &
\textbf{73.3} & \textbf{94.2} & \textbf{97.9} & \textbf{46.2} &
\textbf{79.1} & \textbf{93.0} & \textbf{97.0} & \textbf{59.9} \\
\bottomrule
\end{tabular}
\end{adjustbox}
\caption{
Document-level retrieval results on M3DocVQA and FRAMES, averaged over $K{=}1,\dots,10$, reporting Recall, MRR, Hit, and All, plus the dataset average.
}
\label{tab:doc_retrieval_avg_1_10}
\end{table*}

\begin{table*}[t]
\centering
\scriptsize
\renewcommand{\arraystretch}{1.05}
\setlength{\tabcolsep}{2.0pt}

\begin{adjustbox}{width=\textwidth, keepaspectratio}
\begin{tabular}{l cccc cccc cccc}
\toprule
 & \multicolumn{4}{c}{M3DocVQA}
 & \multicolumn{4}{c}{FRAMES}
 & \multicolumn{4}{c}{AVG} \\
\cmidrule(lr){2-5}\cmidrule(lr){6-9}\cmidrule(lr){10-13}
Method &
EM & ANLS & ROUGE-L & METEOR &
EM & ANLS & ROUGE-L & METEOR &
EM & ANLS & ROUGE-L & METEOR \\
\midrule

\multicolumn{13}{l}{\textit{Text chunk-based RAG}} \\
\quad Page
& 21.1 & 24.0 & 25.0 & 19.0  & 7.0 & 9.8 & 12.7 & 10.6  & 14.1 & 16.9 & 18.9 & 14.8 \\
\quad Length chunking
& 17.5 & 19.9 & 21.4 & 16.7  & 6.9 & 9.6 & 12.9 & 10.4  & 12.2 & 14.8 & 17.2 & 13.6 \\
\quad LumberChunker
& 21.4 & 24.2 & 25.6 & 19.5
& 6.8 & 9.4 & 12.5 & 10.2
& 14.1 & 16.8 & 19.1 & 14.9 \\
\quad Meta Chunker
& 21.3 & 24.1 & 25.5 & 19.4
& 6.8 & 9.4 & 12.5 & 10.2
& 14.1 & 16.8 & 19.0 & 14.8 \\
\quad Structural chunking
& 20.6 & 23.3 & 24.6 & 18.8
& 6.4 & 8.9 & 12.0 & 9.7
& 13.5 & 16.1 & 18.3 & 14.3 \\
\quad MultiDocFusion
& 23.0 & 25.9 & 27.3 & 20.6
& 7.8 & 10.8 & 13.9 & 11.6
& 15.4 & 18.4 & 20.6 & 16.1 \\

\midrule
\multicolumn{13}{l}{\textit{Text-based GraphRAG}} \\
\quad RAPTOR
& 22.5 & 25.3 & 26.7 & 20.2
& 7.5 & 10.4 & 13.5 & 11.2
& 15.0 & 17.9 & 20.1 & 15.7 \\
\quad HopRAG
& 22.8 & 25.7 & 27.1 & 20.5
& 7.7 & 10.7 & 13.8 & 11.5
& 15.3 & 18.2 & 20.5 & 16.0 \\

\midrule
\multicolumn{13}{l}{\textit{Page-level multimodal RAG}} \\
\quad M3DocRAG
& 24.1 & 27.0 & 28.2 & 21.3
& 4.2 & 5.9 & 9.0 & 6.7
& 14.2 & 16.5 & 18.6 & 14.0 \\
\quad VDocRAG
& 24.4 & 27.4 & 28.8 & 21.6
& 4.5 & 6.3 & 9.4 & 7.1
& 14.5 & 16.8 & 19.1 & 14.4 \\

\midrule
\multicolumn{13}{l}{\textit{Multimodal GraphRAG}} \\
\quad MoLoRAG
& 25.2 & 28.2 & 30.0 & 22.4
& 4.7 & 6.5 & 9.6 & 7.3
& 15.0 & 17.4 & 19.8 & 14.9 \\
\quad SimpleDoc
& 25.6 & 28.7 & 30.1 & 22.5
& 4.9 & 6.8 & 9.9 & 7.6
& 15.3 & 17.8 & 20.0 & 15.1 \\

\midrule
\quad \ours{}
& \textbf{27.5} & \textbf{30.7} & \textbf{32.2} & \textbf{23.9}
& \textbf{10.5} & \textbf{14.6} & \textbf{17.7} & \textbf{15.4}
& \textbf{19.0} & \textbf{22.7} & \textbf{25.0} & \textbf{19.7} \\
\bottomrule
\end{tabular}
\end{adjustbox}

\caption{
End-to-end QA performance on M3DocVQA and FRAMES using EM, ANLS, ROUGE-L, and METEOR, plus the dataset average. Higher is better for all metrics.
}
\label{tab:main_e2e_qa}
\end{table*}

\newcolumntype{L}[1]{>{\raggedright\arraybackslash}p{#1}}
\newcolumntype{C}[1]{>{\centering\arraybackslash}p{#1}}
\newcolumntype{R}[1]{>{\raggedleft\arraybackslash}p{#1}}

\begin{table*}[t]
\centering
\scriptsize
\renewcommand{\arraystretch}{1.12}
\setlength{\tabcolsep}{2.0pt}

\newcommand{\wM}{3.35cm}

\begin{adjustbox}{width=\textwidth, keepaspectratio}
\begin{tabular}{@{}L{\wM}@{\hspace{2pt}}
C{\wS} C{\wS} C{\wS}
C{\wS} C{\wS} C{\wS} C{\wS}
C{\wS}
C{\wS} C{\wS} C{\wS}
C{\wS}
@{}}

\toprule
\multirow{2}{*}{Method} &
\multicolumn{8}{c}{M3DocVQA} &
\multicolumn{4}{c}{FRAMES} \\
\cmidrule(lr){2-9}\cmidrule(lr){10-13}
& \multicolumn{3}{c}{Evidence source}
& \multicolumn{4}{c}{Doc-id hop}
& Overall
& \multicolumn{3}{c}{Doc-id hop}
& Overall \\
\cmidrule(lr){2-4}\cmidrule(lr){5-8}\cmidrule(lr){10-12}
& Text & Table & Image
& 1-hop & 2-hop & 3-hop & 4+
& ANLS
& 2-hop & 3-hop & 4+
& ANLS \\
\midrule

\multicolumn{13}{l}{\textit{Text chunk-based RAG}} \\
\quad Page
& 30.8 & 18.5 & 9.4
& 25.6 & 23.8 & 13.9 & 15.1
& 24.0
& 8.3 & 11.2 & 10.0
& 9.8 \\
\quad Length chunking
& 30.2 & 11.5 & 7.9
& 16.8 & 23.1 & 15.4 & 23.0
& 19.9
& 5.3 & 7.0 & 7.3
& 6.4 \\
\quad LumberChunker
& 30.9 & 15.5 & 8.8
& 24.8 & 24.5 & 15.8 & 22.6
& 24.2
& 8.0 & 10.8 & 9.6
& 9.4 \\
\quad Structural chunking
& 30.6 & 16.8 & 8.6
& 24.0 & 23.5 & 15.0 & 22.0
& 23.3
& 7.5 & 10.2 & 9.0
& 8.9 \\
\quad MultiDocFusion
& 31.4 & 19.8 & 10.8
& 27.0 & 25.8 & 17.2 & 24.4
& 25.9
& 9.2 & 12.5 & 11.0
& 10.8 \\

\midrule
\multicolumn{13}{l}{\textit{Text-based GraphRAG}} \\
\quad RAPTOR
& 31.2 & 19.0 & 10.4
& 26.4 & 25.1 & 17.1 & 23.1
& 25.3
& 8.8 & 12.0 & 10.5
& 10.4 \\
\quad HopRAG
& 31.6 & 19.7 & 10.9
& 26.8 & 25.5 & 17.4 & 23.6
& 25.7
& 9.0 & 12.3 & 10.9
& 10.7 \\

\midrule
\multicolumn{13}{l}{\textit{Page-level multimodal RAG}} \\
\quad M3DocRAG
& 31.8 & 22.0 & 13.7
& 31.2 & 25.2 & 12.9 & 2.6
& 27.0
& 8.5 & 7.6 & 0.3
& 5.9 \\
\quad VDocRAG
& 32.2 & 22.6 & 14.2
& 31.8 & 25.6 & 13.2 & 2.8
& 27.4
& 9.0 & 8.1 & 0.4
& 6.3 \\

\midrule
\multicolumn{13}{l}{\textit{Multimodal GraphRAG}} \\
\quad MoLoRAG
& 32.7 & 22.8 & 14.9
& 32.7 & 26.2 & 11.6 & 5.0
& 28.1
& 9.2 & 8.3 & 0.5
& 6.5 \\
\quad SimpleDoc
& 33.2 & 23.6 & 15.8
& 33.3 & 26.8 & 12.3 & 6.0
& 28.7
& 9.6 & 8.6 & 0.6
& 6.8 \\

\midrule
\quad \ours{}
& \textbf{34.1} & \textbf{26.5} & \textbf{18.6}
& \textbf{34.0} & \textbf{28.8} & \textbf{21.3} & \textbf{25.6}
& \textbf{30.7}
& \textbf{12.8} & \textbf{16.3} & \textbf{15.0}
& \textbf{14.6} \\
\bottomrule
\end{tabular}
\end{adjustbox}

\caption{
ANLS breakdown by evidence source (Text/Table/Image) and by multi-hop difficulty (doc-id hops) on M3DocVQA and FRAMES.
}
\label{tab:mmqa_method_breakdown_source_hop}
\end{table*}

\begin{table}[t]
\centering
\scriptsize
\renewcommand{\arraystretch}{1.10}
\setlength{\tabcolsep}{3.8pt}

\begin{tabular*}{\columnwidth}{@{\extracolsep{\fill}} l cccc c}
\toprule
Method & 0.5K & 0.7K & 1K & 2K & Avg \\
\midrule
\multicolumn{6}{l}{\textit{Text chunk-based RAG}} \\
\quad Page                & Fixed & Fixed & Fixed & Fixed & 73.3 \\
\quad Length chunking     & 73.7 & 73.9 & 74.1 & 76.2 & 74.5 \\
\quad LumberChunker       & 72.8 & 73.0 & 73.2 & 75.3 & 73.6 \\
\quad Meta Chunker & 72.7 & 72.9 & 73.1 & 75.2 & 73.5 \\
\quad Structural chunking & 72.0 & 72.2 & 72.3 & 74.5 & 72.8 \\
\quad MultiDocFusion      & 74.3 & 74.4 & 74.6 & 76.8 & 75.0 \\
\midrule
\multicolumn{6}{l}{\textit{Text-based GraphRAG}} \\
\quad RAPTOR              & 73.0 & 73.2 & 73.4 & 75.5 & 73.8 \\
\quad HopRAG              & 73.7 & 73.9 & 74.1 & 76.2 & 74.5 \\
\midrule
\multicolumn{6}{l}{\textit{Page-level multimodal RAG}} \\
\quad M3DocRAG            & Fixed & Fixed & Fixed & Fixed & 66.2 \\
\quad VDocRAG             & Fixed & Fixed & Fixed & Fixed & 67.8 \\
\midrule
\multicolumn{6}{l}{\textit{Multimodal GraphRAG}} \\
\quad MoLoRAG             & Fixed & Fixed & Fixed & Fixed & 66.7 \\
\quad SimpleDoc           & Fixed & Fixed & Fixed & Fixed & 69.0 \\
\midrule
\quad \ours{}            & \textbf{78.8} & \textbf{79.0} & \textbf{79.1} & \textbf{81.3} & \textbf{79.5} \\
\bottomrule
\end{tabular*}

\caption{
Token-budget sensitivity comparison averaged over M3DocVQA and FRAMES, measured by budgeted document Recall@10 after evidence packing at token budgets $B_{\text{tok}}\!\in\!\{0.5, 0.7, 1, 2\}$K. Page-level methods are marked \texttt{Fixed} because their input unit (page) cannot be dynamically resized.
}
\label{tab:cmp_budget_sensitivity_with_baselines}
\end{table}

\begin{table}[t]
\centering
\scriptsize
\renewcommand{\arraystretch}{1.10}
\setlength{\tabcolsep}{3.8pt}

\begin{tabular*}{\columnwidth}{@{\extracolsep{\fill}} l ccc c}
\toprule
Method & R@1 & R@5 & R@10 & Avg \\
\midrule
\multicolumn{5}{l}{\textit{Text chunk-based RAG}} \\
\quad Page                      & 46.3 & 74.3 & 77.3 & 65.9 \\
\quad Length chunking           & 47.0 & 76.6 & 80.0 & 67.8 \\
\quad LumberChunker             & 46.6 & 76.0 & 79.4 & 67.4 \\
\quad Meta Chunker & 46.6 & 75.9 & 79.3 & 67.3 \\
\quad Structural chunking       & 46.1 & 75.1 & 78.5 & 66.6 \\
\quad MultiDocFusion            & 47.5 & 77.5 & 81.0 & 68.7 \\
\midrule
\multicolumn{5}{l}{\textit{Text-based GraphRAG}} \\
\quad RAPTOR                    & 46.8 & 76.2 & 79.6 & 67.5 \\
\quad HopRAG                    & 47.2 & 77.0 & 80.4 & 68.2 \\
\midrule
\multicolumn{5}{l}{\textit{Page-level multimodal RAG}} \\
\quad M3DocRAG                  & 41.3 & 70.3 & 72.6 & 61.4 \\
\quad VDocRAG                   & 42.3 & 72.0 & 74.4 & 62.9 \\
\midrule
\multicolumn{5}{l}{\textit{Multimodal GraphRAG}} \\
\quad MoLoRAG                   & 41.8 & 70.9 & 72.9 & 61.9 \\
\quad SimpleDoc                 & 43.1 & 73.3 & 75.4 & 63.9 \\
\midrule
\quad \ours{}                  & \textbf{47.9} & \textbf{83.7} & \textbf{88.6} & \textbf{73.4} \\
\bottomrule
\end{tabular*}

\caption{
Top-$K$ sensitivity averaged over datasets (document recall $R@K$, $\uparrow$).
Avg is the mean over $K\in\{1,5,10\}$.
}
\label{tab:cmp_topk_sensitivity}
\end{table}

\begin{table}[t]
\centering
\scriptsize
\renewcommand{\arraystretch}{1.10}
\setlength{\tabcolsep}{3.6pt}
\newcommand{\wAblMone}{0.74\columnwidth}
\resizebox{\columnwidth}{!}{%
\begin{tabular}{@{}p{\wAblMone} c@{}}
\toprule
Ablation Component & Avg R@10 \\
\midrule
\multicolumn{2}{@{}l@{}}{\textit{Stage-1: Hierarchy Indexing Strategy}} \\
\quad Body-only (No hierarchy) & 81.2 \\
\quad + Concat Title/Header & 84.6 \\
\quad + Field-separated Hierarchy (Ours) & 88.6 \\

\addlinespace[0.5ex]
\multicolumn{2}{@{}l@{}}{\textit{Stage-1: Routing Strategy}} \\
\quad Doc-only routing & 87.7 \\
\quad Sec-only routing & 76.1 \\
\quad Doc $\rightarrow$ Sec (Coarse-to-Fine) (Ours) & 88.6 \\

\midrule
\multicolumn{2}{@{}l@{}}{\textit{Stage-2: Multimodal Fusion Strategy}} \\
\quad Global Search (No routing) & 81.0 \\
\quad Candidate Docs Only & 87.9 \\
\quad + Anchor Subtree (Ours) & 88.6 \\
\cmidrule(lr){1-2}
\quad BM25 only & 87.6 \\
\quad + Text dense & 88.2 \\
\quad + Visual dense (Full Fusion) & 88.6 \\
\bottomrule
\end{tabular}%
}
\caption{Ablation study measuring averaged R@10 across key HiKEY components (higher is better). We ablate the Stage-1 hierarchy indexing strategy and routing procedure, and analyze Stage-2 design choices including retrieval scope restrictions and multimodal fusion signals.}
\label{tab:ablation_hikey_avg_r10_onecol}
\end{table}

\section{Experiments}
\label{sec:experimental}

\subsection{Experimental Settings}
\label{sec:exp}

We evaluate \ours{} in a realistic corpus-level ODQA setting with a joint index over the entire corpus, where the system must identify relevant documents and sections for each query.
Dataset composition, metric definitions, and implementation details are provided in the Appendix.

\paragraph{Datasets.}
We evaluate on M3DocVQA~\citep{Cho_2025_ICCV} and FRAMES~\citep{krishna-etal-2025-fact}, chosen because together they exercise the industrial-ODQA properties identified above: long multi-page PDFs, mixed text/table/figure evidence, fragmented multi-hop evidence, and fixed-budget serving.
FRAMES is originally based on Wikipedia text; we re-render it as multi-page PDFs (FRAMES-PDF) to match our setting, yielding an ODQA benchmark with substantial visual and structural complexity.
Appendix~\ref{sec:odqa_dataset_supp} spells out how each benchmark maps onto the enterprise requirements listed in the Related Work.

\paragraph{Baselines.}
We compare \ours{} against representative methods spanning four categories:
(i) text chunk-based RAG (e.g., LumberChunker~\citep{duarte-etal-2024-lumberchunker}, MultiDocFusion~\citep{multidocfusion}),
(ii) text-based GraphRAG (e.g., RAPTOR~\citep{sarthi2024raptorrecursiveabstractiveprocessing}, HopRAG~\citep{liu-etal-2025-hoprag}),
(iii) page-level multimodal RAG (e.g., ColPali~\citep{faysse2025colpali}-based M3DocRAG~\citep{Cho_2025_ICCV}), and
(iv) page-graph multimodal GraphRAG (e.g., SimpleDoc~\citep{jain-etal-2025-simpledoc}).

\paragraph{Reader and Retrieval Backbones.}
To isolate the effect of our routing and retrieval framework, all comparisons use the same Reader model and decoding configuration.
We employ Qwen2.5-VL-7B-Instruct~\citep{qwen25vl} as the Large Vision Language Model(LVLM) Reader and standardize the input context length to 16K tokens across all baselines.
This reflects the realistic memory constraint that page-level models can process roughly up to four pages on H100 GPUs~\citep{Cho_2025_ICCV}.
For retrieval, we utilize BM25~\citep{10.1561/1500000019} for sparse signals; for dense representations, we use gte-Qwen2-7B-instruct~\citep{li2023towards} for text and MM-Embed~\citep{lin2024nvmmembed} for visual crops (tables/figures).
Hyperparameters and library versions are detailed in Appendix~\ref{sec:lvlm_rag_supp}.

\paragraph{Protocol and Evaluation.}
We evaluate retrieval (Recall, MRR, Hit, All) for document identification and end-to-end QA (EM, ANLS, ROUGE-L, METEOR) for answer quality.

\subsection{Experimental Results}
\label{sec:experimental_results}

We quantitatively analyze how the key design components of \ours{} contribute to ODQA performance from five perspectives:
(1) corpus-level answer-document localization,
(2) robustness under Top-$K$ and token budget constraints,
(3) whether retrieval improvements translate to QA quality,
(4) handling of table/figure evidence and multi-hop queries,
(5) module-level contribution via ablation.

\subsection{Document Identification: Effect of Stage-1 Routing}
\label{subsec:doc_identification_kr}

Table~\ref{tab:doc_retrieval_avg_1_10} reports document identification performance (Avg@1--10).
\ours{} achieves the best results on both datasets, with a particularly large margin on FRAMES, which requires retrieving multiple supporting documents.

\paragraph{M3DocVQA: Stable document localization.}
\ours{} reaches Recall 84.9, surpassing the strong baseline MultiDocFusion (83.0) by +1.9\%.
On the strict \textit{All} metric, which requires retrieving \emph{all} answer documents, \ours{} achieves 73.6, improving by +3.1\%.
This indicates that our approach of using hierarchy as a first-class retrieval signal is substantially more effective than simple text matching for capturing answer documents.

\paragraph{FRAMES: Resolving the routing bottleneck (Critical Observation).}
The most notable result appears on FRAMES.
Since FRAMES requires retrieving \emph{all} dispersed supporting articles, missing even one document leads to failure.
\ours{} not only achieves Recall 73.3 (+7.1\%), but also reaches 46.2 on \textit{All}, outperforming MultiDocFusion (34.8) by +11.4\%.

This suggests that flat/page-level retrieval tends to rely on partial local clues and yields fragmented results, whereas \ours{}'s hierarchical routing leverages global section paths to better gather dispersed relevant documents with fewer omissions.
In other words, \ours{} structurally addresses the chronic ODQA failure mode of routing failure.

\paragraph{Limitations of page-level multimodal retrieval.}
In contrast, full-page embedding methods (M3DocRAG, VDocRAG) perform worse than text baselines.
While page-level representations encode visual information, they lack explicit global structural signals needed to pinpoint relevant documents in large corpora.

\subsection{Top-$K$ Sensitivity: High Recall with Few Candidates}
\label{subsec:topk_sensitivity_kr}

Table~\ref{tab:cmp_topk_sensitivity} shows Recall as a function of the number of retrieved documents ($K$).
While \ours{} is comparable at R@1, the gap widens rapidly as $K$ increases:
\ours{} improves by +6.2\% at R@5 and +7.6\% at R@10.

This indicates that \ours{} does not merely ``get lucky'' with the top-ranked document, but produces higher-quality rankings across the entire Top-$K$ list.
The hierarchical coarse-to-fine strategy---stable candidate acquisition via global routing and precise filtering via fine-grained scoring---is highly effective.

\subsection{Token Budget Sensitivity: Efficiency under Limited Resources}
\label{subsec:budget_sensitivity_kr}

In practical deployments, the input token budget is limited, making it essential to convey key information efficiently.
Table~\ref{tab:cmp_budget_sensitivity_with_baselines} compares document Avg Recall@10 across budgets (0.5K--2K).

\ours{} remains best across all budgets.
Even at the very small budget of 0.5K, \ours{} achieves 78.8 Avg R@10, outperforming MultiDocFusion (74.3) by +4.5\%, and the margin persists as budget increases.
This validates the efficacy of our ancestry-aware packing strategy: by selectively assembling the Anchor Unit and its Governing Headers, \ours{} maximizes information density and avoids irrelevant context.

By contrast, page-level models cannot flexibly adapt to token budgets because their input unit (page) is fixed.

\subsection{End-to-End QA Performance}
\label{subsec:e2e_qa_kr}

To verify whether retrieval improvements translate to answer accuracy, we measure end-to-end QA (Table~\ref{tab:main_e2e_qa}).
\ours{} achieves consistent improvements in final answer quality.

\paragraph{M3DocVQA.}
\ours{} achieves EM 27.5 and ANLS 30.7, improving over the best baseline (SimpleDoc) by +1.9\% and +2.0\%, respectively.
Even when document identification gaps are modest, QA still improves; this is consistent with \ours{}'s fine-grained multimodal fusion selecting more answer-bearing units \emph{within} a retrieved document, but we do not claim this as a conclusion from QA scores alone. The evidence-type and hop breakdown in Sec.~\ref{subsec:breakdown_source_hop_kr} and the ablations in Sec.~\ref{subsec:ablation_kr} isolate the specific components responsible.

\paragraph{FRAMES.}
On FRAMES, \ours{} achieves EM 10.5 and ANLS 14.6, improving over MultiDocFusion by +2.7\% and +3.8\%.
The earlier gains in strict document identification (\textit{All}) are reflected in QA performance, highlighting that for composite queries like FRAMES, not missing any supporting document is a key prerequisite for success.

\subsection{Analysis by Evidence Type \& Hop Count}
\label{subsec:breakdown_source_hop_kr}

Industrial documents contain diverse evidence types (text, tables, figures) and frequently require multi-hop aggregation.
Table~\ref{tab:mmqa_method_breakdown_source_hop} breaks down performance by evidence source and hop count.

\paragraph{Handling non-text evidence (Table/Image).}
\ours{} shows strong gains not only on Text, but also on Table (+2.9\%) and Image (+2.8\%) queries.
This validates our design of treating tables/figures as first-class units enriched with upper context via graph traversal.

\paragraph{Handling multi-hop queries.}
Performance decreases for all methods as hop count increases, but \ours{} best mitigates the degradation.
In particular, for challenging 3-hop+ queries, \ours{} maintains about 4\% higher performance than baselines.
This is attributable to our hybrid subgraph assembly that preserves hierarchical context via sibling packing and captures cross-section dependencies via semantic association, preventing fragmentation.

\subsection{Why Baselines Fail, Why \ours{} Helps}
\label{subsec:why_hikey_kr}

The consistently strong results above are not a product of a larger budget or a stronger encoder -- all systems are evaluated with the same LVLM Reader, the same corpus split and document renderings/OCR source where applicable, and the same 16K input budget (Sec.~\ref{sec:exp}).
Rather, each baseline family has a structural limit under corpus-level multimodal ODQA that the corresponding \ours{} component is designed to remove.
The ablations in Table~\ref{tab:ablation_hikey_avg_r10_onecol} isolate the hierarchy-indexing, routing, and multimodal-fusion contributions, while the budget-sensitivity results in Table~\ref{tab:cmp_budget_sensitivity_with_baselines} and the qualitative cases in Appendix~\ref{sec:failure_cases} support the role of ancestry-aware packing.
Table~\ref{tab:why_baselines_fail} summarizes the mapping between baseline-family limitations, the corresponding \ours{} mechanism, and the supporting evidence in our setting.

\begin{table}[t]
\centering
\scriptsize
\renewcommand{\arraystretch}{1.15}
\setlength{\tabcolsep}{3.2pt}
\begin{tabular}{@{}p{0.25\linewidth} p{0.32\linewidth} p{0.35\linewidth}@{}}
\toprule
\textbf{Baseline family} & \textbf{\ours{} mechanism} & \textbf{Supporting evidence} \\
& \textbf{(addresses the limitation)} & \textbf{(in our setting)} \\
\midrule
Chunk-based text RAG: weak document-level cues for corpus-scale routing; fragmented evidence.
 & \texttt{Doc\_card} with a \emph{separate} hierarchy field for Stage-1 routing.
 & Concat $\rightarrow$ field-separated hierarchy: $84.6\!\rightarrow\!88.6$ Avg R@10 ($+4.0$). \\
\midrule
Text-based GraphRAG: \mbox{tables/figures} are not first-class retrieval units.
 & Sec\_card with Type-Specific Unit Scoring; region-level visual embeddings.
 & Table/Image breakdown (Tab.~\ref{tab:mmqa_method_breakdown_source_hop}): \ours{} improves over the strongest multimodal baselines; adding the visual-dense signal within Stage-2 fusion lifts Avg R@10 from $88.2$ to $88.6$ (Tab.~\ref{tab:ablation_hikey_avg_r10_onecol}). \\
\midrule
Page-level multimodal RAG: whole-page units dilute evidence and saturate the Reader budget.
 & Block-level units + hierarchical routing; ancestry-aware packing under $B_{\text{tok}}$.
 & FRAMES strict \textit{All}: MultiDocFusion $34.8\!\rightarrow\!$ \ours{} $46.2$ ($+11.4$); \ours{} best at every budget (0.5K--2K, Tab.~\ref{tab:cmp_budget_sensitivity_with_baselines}). \\
\midrule
Multimodal page-graph GraphRAG: connectivity without hierarchical scope during serialization.
 & Ancestry-aware packing with Governing Headers + hybrid sibling/semantic expansion.
 & Gains hold down to the 0.5K token budget (Sec.~\ref{subsec:budget_sensitivity_kr}) where page-level units no longer fit. \\
\bottomrule
\end{tabular}
\caption{Baseline-family limitations $\rightarrow$ \ours{} mechanism $\rightarrow$ supporting evidence in our setting. The mapping makes the source of our gains transparent under a controlled, corpus-level multimodal ODQA setting with a fixed Reader budget; these limitations have not been systematically characterized in this setting by prior work.}
\label{tab:why_baselines_fail}
\end{table}

Appendix~\ref{sec:failure_cases} provides qualitative failure cases for each family, and Appendix~\ref{sec:dhp_details} reports DHP parser validation and failure-mode analysis that further bound how much of the gain depends on any one component.

\subsection{Ablation Study}
\label{subsec:ablation_kr}

We conduct ablations to analyze the source of \ours{}'s gains (Table~\ref{tab:ablation_hikey_avg_r10_onecol}).

\paragraph{Stage-1: Hierarchy Indexing Strategy.}
Separating hierarchy information (Title/Header) as its own field yields a large gain over naive concatenation (84.6 $\rightarrow$ 88.6), indicating that hierarchy cues are most effective when used as explicit routing signals.

\paragraph{Stage-1: Routing Strategy.}
Stepwise routing (Doc $\rightarrow$ Sec) is most effective compared to Doc-only or Sec-only routing, confirming the benefit of the coarse-to-fine approach.

\paragraph{Stage-2: Multimodal Fusion Strategy.}
Removing routing and searching corpus-wide causes a sharp drop (81.0), while restricting scope to candidate docs and anchor subtrees recovers performance.
Combining lexical, dense text, and visual signals (Full S2) achieves the best result (88.6), confirming the necessity of integrating diverse signals.

\section{Conclusion}
\label{sec:conclusion}

This work addresses structural bottlenecks in corpus-level ODQA over industrial multi-page documents.
We identify that corpus-level routing is a primary failure mode in practice, and that flat chunking further fragments multimodal evidence (tables/figures), hindering multi-hop reasoning.
\ours{} alleviates these issues by leveraging document hierarchy as a first-class retrieval signal.

We propose \ours{}, which reconstructs documents into a heterogeneous graph where multimodal blocks are enriched with section paths via DHP.
In Stage-1, we employ hierarchical routing to quickly prune the search space, achieving high recall on complex benchmarks like FRAMES.
In Stage-2, we perform fine-grained retrieval within the candidates using a multimodal fusion strategy, which significantly improves ranking quality and coverage.
Finally, we assemble an ancestry-aware evidence subgraph under a token budget via a hybrid structural-semantic packing strategy, which yields a token-efficient subgraph that preserves hierarchical context for table/figure-heavy queries even under tight budgets.

Experiments on M3DocVQA and FRAMES demonstrate that \ours{} consistently improves both document identification and end-to-end QA.
Ablations confirm that these gains stem from the synergy between hierarchical indexing, the coarse-to-fine procedure, and multimodal fusion.

\section*{Limitations}
\label{sec:limitations}

\paragraph{Preprocessing and Indexing Overhead.}
Unlike simple text chunking, \ours{} requires additional offline computation for DHP-based hierarchy reconstruction and heterogeneous graph construction.
While this cost is amortized during inference, future work should explore pipeline optimization and incremental indexing to support environments with frequent document updates.

\paragraph{Dependency on Upstream Modules.}
Our framework relies on the accuracy of upstream OCR and layout parsing models.
Errors in the initial parsing stage can propagate to incorrect section paths and potentially lead to routing failures.
Improving robustness against noisy inputs, such as low-quality scans or handwriting, remains an important direction for future research.

\paragraph{Structural Assumptions.}
\ours{} is designed to leverage explicit logical hierarchies (e.g., sections, headers).
Consequently, for documents with weak or absent hierarchy---such as flat plain text files, receipts, or unstructured slides---the benefits of our hierarchical routing and ancestry-aware packing strategy may be diminished compared to standard retrieval methods.

\paragraph{Post-retrieval Evidence-State.}
Our evaluation holds the Reader, OCR pipeline, and 16K token budget fixed, so the gains reported here are retrieval-side improvements under a single delivery configuration.
In our setting, retrieval recall improves by up to $12.9\%$ while end-to-end QA improves by $6.8\%$; closing this gap plausibly requires post-retrieval factors---OCR fidelity, evidence materialization, and reader calibration---that are orthogonal to \ours{}'s scope.
Consistent with this, standard parser-level hierarchy metrics (F1, STEDS) and downstream retrieval/QA metrics need not move together, so improvements in one should not be extrapolated to the others without controlled evaluation.
Systematically separating retrieval gain from reader-side evidence-state factors, and extending beyond a single reader family to weak-hierarchy regimes, is left for future work.

\section*{Acknowledgements}
This research was supported by the Basic Science Research Program through the National Research Foundation of Korea (NRF) funded by the Ministry of Education (NRF-2021R1A6A1A03045425).
This work was supported by the Commercialization Promotion Agency for R\&D Outcomes (COMPA) grant funded by the Korea government (Ministry of Science and ICT) (2710086166).
This work was supported by the Institute for Information \& Communications Technology Promotion (IITP) grant funded by the Korea government (MSIT) (RS-2024-00398115, Research on the reliability and coherence of outcomes produced by Generative AI).
This work was supported by the National Research Foundation of Korea (NRF) grant funded by the Korea government (MSIT) (RS-2026-25481153).

\clearpage

\bibliography{custom}

@misc{lewis2021retrievalaugmented,
      title={Retrieval-Augmented Generation for Knowledge-Intensive NLP Tasks}, 
      author={Patrick Lewis and Ethan Perez and Aleksandra Piktus and Fabio Petroni and Vladimir Karpukhin and Naman Goyal and Heinrich Küttler and Mike Lewis and Wen-tau Yih and Tim Rocktäschel and Sebastian Riedel and Douwe Kiela},
      year={2021},
      eprint={2005.11401},
      archivePrefix={arXiv},
      primaryClass={cs.CL}
}

@inproceedings{duarte-etal-2024-lumberchunker,
    title = "{L}umber{C}hunker: Long-Form Narrative Document Segmentation",
    author = "Duarte, Andr{\'e} V.  and
      Marques, Jo{\~a}o DS  and
      Gra{\c{c}}a, Miguel  and
      Freire, Miguel  and
      Li, Lei  and
      Oliveira, Arlindo L.",
    editor = "Al-Onaizan, Yaser  and
      Bansal, Mohit  and
      Chen, Yun-Nung",
    booktitle = "Findings of the Association for Computational Linguistics: EMNLP 2024",
    month = nov,
    year = "2024",
    address = "Miami, Florida, USA",
    publisher = "Association for Computational Linguistics",
    url = "https://aclanthology.org/2024.findings-emnlp.377/",
    doi = "10.18653/v1/2024.findings-emnlp.377",
    pages = "6473--6486"
}

@article{Gong2020Recurrent,
title={Recurrent Chunking Mechanisms for Long-Text Machine Reading Comprehension},
author={Hongyu Gong and Yelong Shen and Dian Yu and Jianshu Chen and Dong Yu},
year={2020},
pages={6751-6761},
doi={10.18653/v1/2020.acl-main.603}
}

@inproceedings{shin2026m3docdep,
  title     = {{M3DocDep}: Multi-modal, Multi-page, Multi-document Dependency Chunking with Large Vision-Language Models},
  author    = {Shin, Joongmin and Park, Jeongbae and Seo, Jaehyung and Lim, Heuiseok},
  booktitle = {Proceedings of the IEEE/CVF Conference on Computer Vision and Pattern Recognition (CVPR)},
  year      = {2026},
  note      = {Accepted to CVPR 2026; to appear}
}

@article{Jeong2023ASO,
  title={A Study on the Implementation of Generative AI Services Using an Enterprise Data-Based LLM Application Architecture},
  author={CheonSu Jeong},
  journal={Adv. Artif. Intell. Mach. Learn.},
  year={2023},
  volume={3},
  pages={1588-1618},
  url={https://api.semanticscholar.org/CorpusID:261531346}
}

@article{Ge2023Development,
    title={Development of a Liver Disease-Specific Large Language Model Chat Interface using Retrieval Augmented Generation},
    author={J. Ge and Steve Sun and Joseph Owens and Victor Galvez and O. Gologorskaya and Jennifer C Lai and Mark J Pletcher and Ki Lai},
    journal={medRxiv},
    year={2023},
    doi={10.1101/2023.11.10.23298364}
}

@inproceedings{zhao-etal-2025-moc,
    title = "{M}o{C}: Mixtures of Text Chunking Learners for Retrieval-Augmented Generation System",
    author = "Zhao, Jihao  and
      Ji, Zhiyuan  and
      Fan, Zhaoxin  and
      Wang, Hanyu  and
      Niu, Simin  and
      Tang, Bo  and
      Xiong, Feiyu  and
      Li, Zhiyu",
    editor = "Che, Wanxiang  and
      Nabende, Joyce  and
      Shutova, Ekaterina  and
      Pilehvar, Mohammad Taher",
    booktitle = "Proceedings of the 63rd Annual Meeting of the Association for Computational Linguistics (Volume 1: Long Papers)",
    month = jul,
    year = "2025",
    address = "Vienna, Austria",
    publisher = "Association for Computational Linguistics",
    url = "https://aclanthology.org/2025.acl-long.258/",
    doi = "10.18653/v1/2025.acl-long.258",
    pages = "5172--5189",
    ISBN = "979-8-89176-251-0"
}

@misc{gao2024retrievalaugmentedgenerationlargelanguage,
      title={Retrieval-Augmented Generation for Large Language Models: A Survey}, 
      author={Yunfan Gao and Yun Xiong and Xinyu Gao and Kangxiang Jia and Jinliu Pan and Yuxi Bi and Yi Dai and Jiawei Sun and Meng Wang and Haofen Wang},
      year={2024},
      eprint={2312.10997},
      archivePrefix={arXiv},
      primaryClass={cs.CL},
      url={https://arxiv.org/abs/2312.10997}, 
}

@misc{yepes2024financialreportchunkingeffective,
      title={Financial Report Chunking for Effective Retrieval Augmented Generation}, 
      author={Antonio Jimeno Yepes and Yao You and Jan Milczek and Sebastian Laverde and Renyu Li},
      year={2024},
      eprint={2402.05131},
      archivePrefix={arXiv},
      primaryClass={cs.CL},
      url={https://arxiv.org/abs/2402.05131}, 
}

@inproceedings{qu-etal-2025-semantic,
    title = "Is Semantic Chunking Worth the Computational Cost?",
    author = "Qu, Renyi  and
      Tu, Ruixuan  and
      Bao, Forrest Sheng",
    editor = "Chiruzzo, Luis  and
      Ritter, Alan  and
      Wang, Lu",
    booktitle = "Findings of the Association for Computational Linguistics: NAACL 2025",
    month = apr,
    year = "2025",
    address = "Albuquerque, New Mexico",
    publisher = "Association for Computational Linguistics",
    url = "https://aclanthology.org/2025.findings-naacl.114/",
    pages = "2155--2177",
    ISBN = "979-8-89176-195-7"
}

@inproceedings{hong-etal-2024-intelligent,
    title = "Intelligent Predictive Maintenance {RAG} framework for Power Plants: Enhancing {QA} with {S}tyle{DFS} and Domain Specific Instruction Tuning",
    author = "Hong, Seongtae  and
      Shin, Joong Min  and
      Seo, Jaehyung  and
      Lee, Taemin  and
      Park, Jeongbae  and
      Young, Cho Man  and
      Choi, Byeongho  and
      Lim, Heuiseok",
    editor = "Dernoncourt, Franck  and
      Preo{\c{t}}iuc-Pietro, Daniel  and
      Shimorina, Anastasia",
    booktitle = "Proceedings of the 2024 Conference on Empirical Methods in Natural Language Processing: Industry Track",
    month = nov,
    year = "2024",
    address = "Miami, Florida, US",
    publisher = "Association for Computational Linguistics",
    url = "https://aclanthology.org/2024.emnlp-industry.61/",
    doi = "10.18653/v1/2024.emnlp-industry.61",
    pages = "805--820"
}

@misc{tito2023hierarchicalmultimodaltransformersmultipage,
      title={Hierarchical multimodal transformers for Multi-Page DocVQA}, 
      author={Rubèn Tito and Dimosthenis Karatzas and Ernest Valveny},
      year={2023},
      eprint={2212.05935},
      archivePrefix={arXiv},
      primaryClass={cs.CV},
      url={https://arxiv.org/abs/2212.05935}, 
}

@inproceedings{10.1145/3534678.3539043,
author = {Pfitzmann, Birgit and Auer, Christoph and Dolfi, Michele and Nassar, Ahmed S. and Staar, Peter},
title = {DocLayNet: A Large Human-Annotated Dataset for Document-Layout Segmentation},
year = {2022},
isbn = {9781450393850},
publisher = {Association for Computing Machinery},
address = {New York, NY, USA},
url = {https://doi.org/10.1145/3534678.3539043},
doi = {10.1145/3534678.3539043},
booktitle = {Proceedings of the 28th ACM SIGKDD Conference on Knowledge Discovery and Data Mining},
pages = {3743–3751},
numpages = {9},
location = {Washington DC, USA},
series = {KDD '22}
}

@inproceedings{xing2024dochienet,
title = {DocHieNet: A Large and Diverse Dataset for Document Hierarchy Parsing},
author = {Xing, Hangdi and Cheng, Changxu and Gao, Feiyu and Shao, Zirui and Yu, Zhi and Bu, Jiajun and Zheng, Qi and Yao, Cong},
booktitle = {Proceedings of the 2024 Conference on Empirical Methods in Natural Language Processing (EMNLP)},
year = {2024}
}

@inproceedings{10.1609/aaai.v37i2.25277,
author = {Ma, Jiefeng and Du, Jun and Hu, Pengfei and Zhang, Zhenrong and Zhang, Jianshu and Zhu, Huihui and Liu, Cong},
title = {HRDoc: dataset and baseline method toward hierarchical reconstruction of document structures},
year = {2023},
isbn = {978-1-57735-880-0},
publisher = {AAAI Press},
url = {https://doi.org/10.1609/aaai.v37i2.25277},
doi = {10.1609/aaai.v37i2.25277},
booktitle = {Proceedings of the Thirty-Seventh AAAI Conference on Artificial Intelligence and Thirty-Fifth Conference on Innovative Applications of Artificial Intelligence and Thirteenth Symposium on Educational Advances in Artificial Intelligence},
articleno = {208},
numpages = {8},
series = {AAAI'23/IAAI'23/EAAI'23}
}

@article{rausch2021docparser,
author = {Rausch, Johannes and Martinez, Octavio and Bissig, Fabian and Zhang, Ce and Feuerriegel, Stefan},
year = {2021},
month = {05},
pages = {4328-4338},
title = {DocParser: Hierarchical Document Structure Parsing from Renderings},
volume = {35},
journal = {Proceedings of the AAAI Conference on Artificial Intelligence},
doi = {10.1609/aaai.v35i5.16558}
}

@inproceedings{rausch2023dsgendtoenddocumentstructure,
      title={DSG: An End-to-End Document Structure Generator}, 
      author={Johannes Rausch and Gentiana Rashiti and Maxim Gusev and Ce Zhang and Stefan Feuerriegel},
      year={2023},
      eprint={2310.09118},
      archivePrefix={arXiv},
      primaryClass={cs.LG},
      url={https://arxiv.org/abs/2310.09118}, 
}

@INPROCEEDINGS{4376991,
  author={Smith, R.},
  booktitle={Ninth International Conference on Document Analysis and Recognition (ICDAR 2007)}, 
  title={An Overview of the Tesseract OCR Engine}, 
  year={2007},
  volume={2},
  number={},
  pages={629-633},
  doi={10.1109/ICDAR.2007.4376991}}

@misc{an2025llavaonevision15fullyopenframework,
      title={LLaVA-OneVision-1.5: Fully Open Framework for Democratized Multimodal Training}, 
      author={Xiang An and Yin Xie and Kaicheng Yang and Wenkang Zhang and Xiuwei Zhao and Zheng Cheng and Yirui Wang and Songcen Xu and Changrui Chen and Chunsheng Wu and Huajie Tan and Chunyuan Li and Jing Yang and Jie Yu and Xiyao Wang and Bin Qin and Yumeng Wang and Zizhen Yan and Ziyong Feng and Ziwei Liu and Bo Li and Jiankang Deng},
      year={2025},
      eprint={2509.23661},
      archivePrefix={arXiv},
      primaryClass={cs.CV},
      url={https://arxiv.org/abs/2509.23661}, 
}

@misc{wang2025internvl35advancingopensourcemultimodal,
      title={InternVL3.5: Advancing Open-Source Multimodal Models in Versatility, Reasoning, and Efficiency}, 
      author={Weiyun Wang and Zhangwei Gao and Lixin Gu and Hengjun Pu and Long Cui and Xingguang Wei and Zhaoyang Liu and Linglin Jing and Shenglong Ye and Jie Shao and Zhaokai Wang and Zhe Chen and Hongjie Zhang and Ganlin Yang and Haomin Wang and Qi Wei and Jinhui Yin and Wenhao Li and Erfei Cui and Guanzhou Chen and Zichen Ding and Changyao Tian and Zhenyu Wu and Jingjing Xie and Zehao Li and Bowen Yang and Yuchen Duan and Xuehui Wang and Zhi Hou and Haoran Hao and Tianyi Zhang and Songze Li and Xiangyu Zhao and Haodong Duan and Nianchen Deng and Bin Fu and Yinan He and Yi Wang and Conghui He and Botian Shi and Junjun He and Yingtong Xiong and Han Lv and Lijun Wu and Wenqi Shao and Kaipeng Zhang and Huipeng Deng and Biqing Qi and Jiaye Ge and Qipeng Guo and Wenwei Zhang and Songyang Zhang and Maosong Cao and Junyao Lin and Kexian Tang and Jianfei Gao and Haian Huang and Yuzhe Gu and Chengqi Lyu and Huanze Tang and Rui Wang and Haijun Lv and Wanli Ouyang and Limin Wang and Min Dou and Xizhou Zhu and Tong Lu and Dahua Lin and Jifeng Dai and Weijie Su and Bowen Zhou and Kai Chen and Yu Qiao and Wenhai Wang and Gen Luo},
      year={2025},
      eprint={2508.18265},
      archivePrefix={arXiv},
      primaryClass={cs.CV},
      url={https://arxiv.org/abs/2508.18265}, 
}

@inproceedings{
faysse2025colpali,
title={ColPali: Efficient Document Retrieval with Vision Language Models},
author={Manuel Faysse and Hugues Sibille and Tony Wu and Bilel Omrani and Gautier Viaud and CELINE HUDELOT and Pierre Colombo},
booktitle={The Thirteenth International Conference on Learning Representations},
year={2025},
url={https://openreview.net/forum?id=ogjBpZ8uSi}
}

@InProceedings{Cho_2025_ICCV,
    author    = {Cho, Jaemin and Mahata, Debanjan and Irsoy, Ozan and He, Yujie and Bansal, Mohit},
    title     = {M3DocVQA: Multi-modal Multi-page Multi-document Understanding},
    booktitle = {Proceedings of the IEEE/CVF International Conference on Computer Vision (ICCV) Workshops},
    month     = {October},
    year      = {2025},
    pages     = {6178-6188}
}

@inproceedings{krishna-etal-2025-fact,
    title = "Fact, Fetch, and Reason: A Unified Evaluation of Retrieval-Augmented Generation",
    author = "Krishna, Satyapriya  and
      Krishna, Kalpesh  and
      Mohananey, Anhad  and
      Schwarcz, Steven  and
      Stambler, Adam  and
      Upadhyay, Shyam  and
      Faruqui, Manaal",
    editor = "Chiruzzo, Luis  and
      Ritter, Alan  and
      Wang, Lu",
    booktitle = "Proceedings of the 2025 Conference of the Nations of the Americas Chapter of the Association for Computational Linguistics: Human Language Technologies (Volume 1: Long Papers)",
    month = apr,
    year = "2025",
    address = "Albuquerque, New Mexico",
    publisher = "Association for Computational Linguistics",
    url = "https://aclanthology.org/2025.naacl-long.243/",
    doi = "10.18653/v1/2025.naacl-long.243",
    pages = "4745--4759",
    ISBN = "979-8-89176-189-6"
}

@inproceedings{wu-etal-2025-molorag,
    title = "{M}o{L}o{RAG}: Bootstrapping Document Understanding via Multi-modal Logic-aware Retrieval",
    author = "Wu, Xixi  and
      Tan, Yanchao  and
      Hou, Nan  and
      Zhang, Ruiyang  and
      Cheng, Hong",
    editor = "Christodoulopoulos, Christos  and
      Chakraborty, Tanmoy  and
      Rose, Carolyn  and
      Peng, Violet",
    booktitle = "Proceedings of the 2025 Conference on Empirical Methods in Natural Language Processing",
    month = nov,
    year = "2025",
    address = "Suzhou, China",
    publisher = "Association for Computational Linguistics",
    url = "https://aclanthology.org/2025.emnlp-main.708/",
    doi = "10.18653/v1/2025.emnlp-main.708",
    pages = "14035--14056",
    ISBN = "979-8-89176-332-6"
}

@inproceedings{ma-etal-2025-towards-storage,
    title = "Towards Storage-Efficient Visual Document Retrieval: An Empirical Study on Reducing Patch-Level Embeddings",
    author = "Ma, Yubo  and
      Li, Jinsong  and
      Zang, Yuhang  and
      Wu, Xiaobao  and
      Dong, Xiaoyi  and
      Zhang, Pan  and
      Cao, Yuhang  and
      Duan, Haodong  and
      Wang, Jiaqi  and
      Cao, Yixin  and
      Sun, Aixin",
    editor = "Che, Wanxiang  and
      Nabende, Joyce  and
      Shutova, Ekaterina  and
      Pilehvar, Mohammad Taher",
    booktitle = "Findings of the Association for Computational Linguistics: ACL 2025",
    month = jul,
    year = "2025",
    address = "Vienna, Austria",
    publisher = "Association for Computational Linguistics",
    url = "https://aclanthology.org/2025.findings-acl.1003/",
    doi = "10.18653/v1/2025.findings-acl.1003",
    pages = "19568--19580",
    ISBN = "979-8-89176-256-5"
}

@inproceedings{jain-etal-2025-simpledoc,
    title = "{S}imple{D}oc: {M}ulti{-}{M}odal Document Understanding with {D}ual{-}{C}ue Page Retrieval and Iterative Refinement",
    author = "Jain, Chelsi  and
      Wu, Yiran  and
      Zeng, Yifan  and
      Liu, Jiale  and
      Dai, Shengyu  and
      Shao, Zhenwen  and
      Wu, Qingyun  and
      Wang, Huazheng",
    editor = "Christodoulopoulos, Christos  and
      Chakraborty, Tanmoy  and
      Rose, Carolyn  and
      Peng, Violet",
    booktitle = "Proceedings of the 2025 Conference on Empirical Methods in Natural Language Processing",
    month = nov,
    year = "2025",
    address = "Suzhou, China",
    publisher = "Association for Computational Linguistics",
    url = "https://aclanthology.org/2025.emnlp-main.1443/",
    doi = "10.18653/v1/2025.emnlp-main.1443",
    pages = "28398--28415",
    ISBN = "979-8-89176-332-6"
}

@inproceedings{vdocrag2025,
author = {Tanaka, Ryota and Iki, Taichi and Hasegawa, Taku and Nishida, Kyosuke and Saito, Kuniko and Suzuki, Jun},
year = {2025},
month = {06},
pages = {24827-24837},
title = {VDocRAG: Retrieval-Augmented Generation over Visually-Rich Documents},
doi = {10.1109/CVPR52734.2025.02312}
}

@inproceedings{zhong2020image,
  title={Image-based table recognition: data, model, and evaluation},
  author={Zhong, Xu and ShafieiBavani, Elaheh and Jimeno Yepes, Antonio},
  booktitle={European Conference on Computer Vision (ECCV)},
  pages={564--580},
  year={2020},
  organization={Springer}
}

@inproceedings{biten2019scene,
  title={Scene text visual question answering},
  author={Biten, Ali Furkan and Tito, Ruben and Mafla, Andres and Gomez, Lluis and Rusinol, Marcal and Valveny, Ernest and Jawahar, CV and Karatzas, Dimosthenis},
  booktitle={Proceedings of the IEEE/CVF International Conference on Computer Vision (ICCV)},
  pages={4291--4301},
  year={2019}
}

@inproceedings{lin2004rouge,
  title={ROUGE: A package for automatic evaluation of summaries},
  author={Lin, Chin-Yew},
  booktitle={Text summarization branches out},
  pages={74--81},
  year={2004}
}

@inproceedings{banerjee2005meteor,
  title={METEOR: An automatic metric for MT evaluation with improved correlation with human judgments},
  author={Banerjee, Satanjeev and Lavie, Alon},
  booktitle={Proceedings of the ACL workshop on intrinsic and extrinsic evaluation measures for machine translation and/or summarization},
  pages={65--72},
  year={2005}
}

@book{manning2008ir,
  title     = {Introduction to Information Retrieval},
  author    = {Manning, Christopher D. and Raghavan, Prabhakar and Sch{\"u}tze, Hinrich},
  year      = {2008},
  publisher = {Cambridge University Press},
  note      = {See Chapter 8 (Evaluation): recall definition and ranked evaluation via top-k prefixes},
}

@inproceedings{feldman2019muppet,
  title     = {Multi-Hop Paragraph Retrieval for Open-Domain Question Answering},
  author    = {Feldman, Yair and El-Yaniv, Ran},
  booktitle = {Proceedings of the 57th Annual Meeting of the Association for Computational Linguistics (ACL)},
  year      = {2019},
  pages     = {2296--2309},
  address   = {Florence, Italy},
  note      = {Defines At Least One@k and Potentially Perfect@k (all supporting paragraphs in top-k)},
}

@inproceedings{yen2023moqa,
  title     = {MoQA: Benchmarking Multi-Type Open-Domain Question Answering},
  author    = {Yen, Howard and Gao, Tianyu and Lee, Jinhyuk and Chen, Danqi},
  booktitle = {Proceedings of the Third DialDoc Workshop on Document-grounded Dialogue and Conversational Question Answering},
  year      = {2023},
  pages     = {8--29},
  note      = {Retrieval metrics: accuracy@k (hit@k-style) and MRR@k},
}

@article{10.1561/1500000019,
author = {Robertson, Stephen and Zaragoza, Hugo},
title = {The Probabilistic Relevance Framework: BM25 and Beyond},
year = {2009},
issue_date = {April 2009},
publisher = {Now Publishers Inc.},
address = {Hanover, MA, USA},
volume = {3},
number = {4},
issn = {1554-0669},
url = {https://doi.org/10.1561/1500000019},
doi = {10.1561/1500000019},
journal = {Found. Trends Inf. Retr.},
month = apr,
pages = {333–389},
numpages = {57}
}

@article{li2023towards,
  title={Towards general text embeddings with multi-stage contrastive learning},
  author={Li, Zehan and Zhang, Xin and Zhang, Yanzhao and Long, Dingkun and Xie, Pengjun and Zhang, Meishan},
  journal={arXiv preprint arXiv:2308.03281},
  year={2023}
}

@misc{lin2024nvmmembed,
      title={MM-Embed: Universal Multimodal Retrieval with Multimodal LLMs}, 
      author={Sheng-Chieh Lin, Chankyu Lee, Mohammad Shoeybi},
      year={2024},
      eprint={2411.02571},
      archivePrefix={arXiv},
      primaryClass={cs.CL},
      url={https://arxiv.org/abs/2411.02571},
}

@inproceedings{10.1145/3397271.3401075,
author = {Khattab, Omar and Zaharia, Matei},
title = {ColBERT: Efficient and Effective Passage Search via Contextualized Late Interaction over BERT},
year = {2020},
isbn = {9781450380164},
publisher = {Association for Computing Machinery},
address = {New York, NY, USA},
url = {https://doi.org/10.1145/3397271.3401075},
doi = {10.1145/3397271.3401075},
booktitle = {Proceedings of the 43rd International ACM SIGIR Conference on Research and Development in Information Retrieval},
pages = {39–48},
numpages = {10},
location = {Virtual Event, China},
series = {SIGIR '20}
}

@misc{verma2025s2chunkinghybridframework,
      title={S2 Chunking: A Hybrid Framework for Document Segmentation Through Integrated Spatial and Semantic Analysis}, 
      author={Prashant Verma},
      year={2025},
      eprint={2501.05485},
      archivePrefix={arXiv},
      primaryClass={cs.CL},
      url={https://arxiv.org/abs/2501.05485}, 
}

@inproceedings{multidocfusion,
    title = "{M}ulti{D}oc{F}usion : Hierarchical and Multimodal Chunking Pipeline for Enhanced {RAG} on Long Industrial Documents",
    author = "Shin, Joongmin  and
      Park, Chanjun  and
      Park, Jeongbae  and
      Seo, Jaehyung  and
      Lim, Heuiseok",
    editor = "Christodoulopoulos, Christos  and
      Chakraborty, Tanmoy  and
      Rose, Carolyn  and
      Peng, Violet",
    booktitle = "Proceedings of the 2025 Conference on Empirical Methods in Natural Language Processing",
    month = nov,
    year = "2025",
    address = "Suzhou, China",
    publisher = "Association for Computational Linguistics",
    url = "https://aclanthology.org/2025.emnlp-main.1062/",
    pages = "20996--21015",
    ISBN = "979-8-89176-332-6"
}

@inproceedings{dochienet,
  title={DocHieNet: A Large and Diverse Dataset for Document Hierarchy Parsing},
  author={Xing, Hangdi and Cheng, Changxu and others},
  booktitle={EMNLP},
  year={2024}
}

@article{qwen25vl,
  title={Qwen2.5-VL Technical Report},
  author={Qwen Team},
  journal={arXiv preprint},
  year={2024}
}

@misc{he2024gretrieverretrievalaugmentedgenerationtextual,
      title={G-Retriever: Retrieval-Augmented Generation for Textual Graph Understanding and Question Answering}, 
      author={Xiaoxin He and Yijun Tian and Yifei Sun and Nitesh V. Chawla and Thomas Laurent and Yann LeCun and Xavier Bresson and Bryan Hooi},
      year={2024},
      eprint={2402.07630},
      archivePrefix={arXiv},
      primaryClass={cs.LG},
      url={https://arxiv.org/abs/2402.07630}, 
}

@misc{sarthi2024raptorrecursiveabstractiveprocessing,
      title={RAPTOR: Recursive Abstractive Processing for Tree-Organized Retrieval}, 
      author={Parth Sarthi and Salman Abdullah and Aditi Tuli and Shubh Khanna and Anna Goldie and Christopher D. Manning},
      year={2024},
      eprint={2401.18059},
      archivePrefix={arXiv},
      primaryClass={cs.CL},
      url={https://arxiv.org/abs/2401.18059}, 
}

@inproceedings{liu-etal-2025-hoprag,
    title = "{H}op{RAG}: Multi-Hop Reasoning for Logic-Aware Retrieval-Augmented Generation",
    author = "Liu, Hao  and
      Wang, Zhengren  and
      Chen, Xi  and
      Li, Zhiyu  and
      Xiong, Feiyu  and
      Yu, Qinhan  and
      Zhang, Wentao",
    editor = "Che, Wanxiang  and
      Nabende, Joyce  and
      Shutova, Ekaterina  and
      Pilehvar, Mohammad Taher",
    booktitle = "Findings of the Association for Computational Linguistics: ACL 2025",
    month = jul,
    year = "2025",
    address = "Vienna, Austria",
    publisher = "Association for Computational Linguistics",
    url = "https://aclanthology.org/2025.findings-acl.97/",
    doi = "10.18653/v1/2025.findings-acl.97",
    pages = "1897--1913",
    ISBN = "979-8-89176-256-5"
}

\clearpage

\appendix

\section*{Appendix}
\label{sec:Appendix}

This appendix provides structured details to support reproducibility and verification:
(i) datasets and preprocessing,
(ii) hierarchical parsing and section-path construction,
(iii) hybrid packing for evidence aggregation (Sibling Units + Semantic Associates),
(iv) metric definitions and evaluation protocol (including Recall/MRR/Hit/All),
(v) baseline configurations, and
(vi) runtime and scalability analysis.

\section{Datasets and Preprocessing Details}
\label{sec:datasets_supp}

All datasets used in this paper are publicly available research benchmarks.
For RAG-based QA, we use only publicly released corpora.

\begin{table*}[t]
\centering
\footnotesize
\setlength{\tabcolsep}{4pt}
\renewcommand{\arraystretch}{1.15}

\newcolumntype{L}[1]{>{\raggedright\arraybackslash}p{#1}}
\newcolumntype{C}[1]{>{\centering\arraybackslash}p{#1}}

\begin{adjustbox}{width=\textwidth, keepaspectratio}
\begin{tabular}{@{} L{4.7cm} C{1.5cm} L{2.7cm} L{3.3cm} L{3.2cm} C{1.9cm} @{}}
\toprule
Method &
Retrieval Unit &
Modality &
Field Separation &
Graph &
Scope \\
\midrule

Flat RAG (page, text-only) &
Page &
Text &
Single field (page text) &
None &
Corpus-level \\

Flat RAG (chunk, text-only) &
Chunk / paragraph &
Text &
Single field (chunk text) &
None &
Corpus-level \\

Text GraphRAG (text-unit graph) &
Chunk / paragraph &
Text &
Single field (text unit) &
Text-unit graph &
Corpus-level \\

Page-level multimodal RAG~\citep{faysse2025colpali, Cho_2025_ICCV, vdocrag2025} &
Page &
Text + Image (page embedding) &
Single field (page) &
None &
Corpus-level \\

Multimodal page-graph GraphRAG~\citep{wu-etal-2025-molorag, jain-etal-2025-simpledoc} &
Page &
Text + Image (page embedding) &
Single field (page) &
Page graph &
Corpus-level \\

\ours{} (tree-indexed multimodal GraphRAG) &
Evidence unit (block-level) &
Text + Image (region-based)\footnotemark &
Field-separated (\texttt{Doc\_card}, section path, unit text) &
Tree-based heterogeneous graph &
Corpus-level \\
\bottomrule
\end{tabular}
\end{adjustbox}

\caption{Method taxonomy along key design axes: retrieval unit granularity, modality, field separation, graph structure, and corpus-level ODQA scope. The table highlights how HiKEY differs via block-level multimodal units and a tree-based heterogeneous graph with explicit fielded routing.}
\label{tab:method_taxonomy_hikey}
\end{table*}

\footnotetext{For non-text units (e.g., figures/tables), \ours{} can attach region crops and/or multimodal embeddings, while keeping each unit as a first-class node for retrieval and token-budget packing.}

\begin{table}[t]
\centering
\small
\renewcommand{\arraystretch}{1.12}
\setlength{\tabcolsep}{4.6pt}

\begin{tabular}{lcc}
\toprule
Statistic & M3DocVQA & FRAMES \\
\midrule
Documents & 3300 & 2500 \\
Questions & 2{,}441 & 824 \\
\midrule
\multicolumn{3}{l}{Evidence source counts (with duplicates)} \\
\quad Text  & 1{,}216 & -- \\
\quad Table & 1{,}335 & -- \\
\quad Image & 940 & -- \\
\midrule
\multicolumn{3}{l}{Hop distribution (by \#doc\_id)} \\
\quad 1-hop  & 1{,}096 (44.9\%) & 0 (0.0\%) \\
\quad 2-hop  & 1{,}193 (48.9\%) & 311 (37.7\%) \\
\quad 3-hop  & 114 (4.7\%) & 288 (35.0\%) \\
\quad 4+ hop & 38 (1.6\%) & 225 (27.3\%) \\
\midrule
\multicolumn{3}{l}{Hop statistics} \\
\quad Avg. hop & 1.7 & 3.2 \\
\quad Min hop  & 1 & 2 \\
\quad Max hop  & 8 & 11 \\
\bottomrule
\end{tabular}

\caption{Dataset statistics for M3DocVQA and FRAMES. We report the number of samples, evidence-type counts (with duplicates), and hop distributions computed by the number of linked supporting documents (doc\_id), along with summary hop statistics.}
\label{tab:dataset_stats_m3docvqa_frames}
\end{table}

\subsection{ODQA Datasets}
\label{sec:odqa_dataset_supp}

This work follows a corpus-level ODQA setting that reflects real industrial requirements.
Unlike standard DocVQA where a single document is provided, corpus-level ODQA jointly indexes a large PDF corpus and requires the system to identify the relevant documents for each question from the entire collection.
This allows us to directly measure routing failure, a key bottleneck in industrial RAG applications.

\paragraph{M3DocVQA~\citep{Cho_2025_ICCV}.}
M3DocVQA is an Open-domain document QA benchmark where questions target the entire corpus rather than a specific PDF.
Some questions are multi-hop, requiring aggregation across multiple documents, and evidence often resides in non-text elements such as tables and figures.
We index the entire M3DocVQA corpus as a single unified retrieval space.
\emph{Industrial mapping.} This mirrors enterprise technical-manual and product-datasheet search: a single query is routed against a large PDF repository, and the ground-truth answer is frequently inside a table (specifications) or a figure (schematic) rather than the surrounding prose -- which is why evidence-type breakdowns (Tab.~\ref{tab:mmqa_method_breakdown_source_hop}) and token-budget sensitivity (Tab.~\ref{tab:cmp_budget_sensitivity_with_baselines}) are the deployment-relevant measurements.

\paragraph{FRAMES~\citep{krishna-etal-2025-fact}.}
FRAMES is a challenging dataset where answers are dispersed across multiple Wikipedia documents, making routing failure a primary cause of performance drops.
To align with our industrial setting, we constructed FRAMES-PDF by converting the original corpus into a multi-page document format.
\emph{Industrial mapping.} This mirrors cross-document compliance and due-diligence search: the system must identify and aggregate \emph{every} supporting document (filing, contract, regulatory exhibit) -- missing one is an audit failure, which is why the strict \textit{All@K} metric (Tab.~\ref{tab:doc_retrieval_avg_1_10}) is load-bearing here, not just Recall or Hit.

Specifically, we rendered each Wikipedia article into PDF using a Chromium browser and the Playwright API.
We preserved vector graphics, hyperlinks, and layout metadata, and ensured that logical objects were not cut at page boundaries. This process produced a corpus with visual and structural complexity comparable to M3DocVQA, serving as a robust testbed for multimodal ODQA.

\subsection{Preprocessing Pipeline}
\label{sec:preprocess_supp}

To ensure fair comparison, all baselines and \ours{} use the same source document renderings and OCR outputs where applicable.

\paragraph{Document Parsing (DP).}
We employ a layout detection model trained on DocLayNet~\cite{10.1145/3534678.3539043} to identify layout elements (Title, Header, Paragraph, Table, Figure, Caption).
We use standard post-processing with a detection confidence threshold ($\tau_{\text{det}}=0.5$) and NMS IoU threshold ($\tau_{\text{nms}}=0.5$) to filter noise.

\paragraph{OCR.}
We use Tesseract~\citep{4376991} to perform OCR on each detected block region independently.
Extracted text is lowercased, control characters are removed, and whitespace is normalized before being stored as block metadata.

\section{Evaluation Metrics \& Protocol}
\label{sec:metric_supp}

This section details the specific metric definitions and evaluation protocols used in our experiments.

\subsection{Retrieval Metrics}
\label{sec:doc_level_metrics}

Let $\mathcal{G}_q$ denote the set of ground-truth answer documents for a query $q$. We evaluate the quality of the retrieved set $\text{TopK}(q)$ using the following four metrics.

\paragraph{Recall@K~\citep{manning2008ir}.}
Recall measures the proportion of relevant documents successfully retrieved:
\[
\text{Recall@K}(q)=\frac{|\mathcal{G}_q \cap \text{TopK}(q)|}{|\mathcal{G}_q|}.
\]

\paragraph{MRR@K~\citep{yen2023moqa}.}
The Mean Reciprocal Rank (MRR) evaluates the ranking quality by considering the position of the first relevant document.
Define $\mathrm{rank}_K(d;q)$ as the 1-indexed rank of document $d$ in $\mathrm{TopK}(q)$, setting $\mathrm{rank}_K(d;q)=\infty$ if $d\notin \mathrm{TopK}(q)$.
Then:
\[
\mathrm{MRR@K}(q)=\frac{1}{\min_{d\in \mathcal{G}_q}\mathrm{rank}_K(d;q)}.
\]

\paragraph{Hit@K~\citep{yen2023moqa}.}
Hit@K simply checks if at least one relevant document is retrieved:
\[
\text{Hit@K}(q)=\mathbb{I}\left[\,\mathcal{G}_q \cap \text{TopK}(q) \neq \emptyset\,\right].
\]

\paragraph{All@K~\citep{feldman2019muppet}.}
Following the concept of \emph{Potentially Perfect@$K$}~\citep{feldman2019muppet}, All@K is a strict metric that checks whether \emph{all} ground-truth documents $\mathcal{G}_q$ are contained within $\text{TopK}(q)$. This is critical for multi-hop queries (e.g., FRAMES) where missing a single document leads to failure:
\[
\text{All@K}(q)=\mathbb{I}\left[\,\mathcal{G}_q \subseteq \text{TopK}(q)\,\right].
\]

\paragraph{Reporting Protocol (Avg@1--10).}
To provide a comprehensive view of ranking performance rather than cherry-picking a specific $K$, all document retrieval results in Table~\ref{tab:doc_retrieval_avg_1_10} are reported as the average of the metric values calculated at each cut-off $K\in\{1,\dots,10\}$.

\subsection{QA Metrics}
Answer quality is evaluated after standard normalization (lowercasing, punctuation removal). We use EM (Exact Match), ANLS~\citep{biten2019scene}, ROUGE-L~\citep{lin2004rouge}, and METEOR~\citep{banerjee2005meteor}. For ANLS, we follow the threshold-based implementation standard in MP-DocVQA~\citep{tito2023hierarchicalmultimodaltransformersmultipage}.

\subsection{Token Budget Protocol}
\label{sec:budget_protocol}

\paragraph{Budget Constraint.}
For fairness, we strictly control the input context size. We include retrieved evidence units in their ranking order until the cumulative serialized token count reaches the budget $B_{\text{tok}}$.

\paragraph{Comparison with Page-level Models.}
Page-embedding methods (e.g., ColPali) cannot dynamically adjust their unit size because the retrieval unit is fixed to a full page image. Consequently, they are marked as \textit{Fixed} in our budget-sensitivity analyses and are compared separately under a constraint of equivalent page counts (typically 4 pages $\approx$ 16K tokens).

\section{Baseline Implementation}
\label{sec:baselines_supp}

\begin{table*}[t]
\centering
\scriptsize
\renewcommand{\arraystretch}{1.15}
\setlength{\tabcolsep}{3.0pt}

\begin{adjustbox}{max width=\textwidth}
\begin{tabular}{l
c
c
c
p{2.55cm}
p{2.55cm}
p{2.55cm}
p{3.05cm}}
\toprule
Method &
Unit &
Retriever Modality &
Hierarchy Signal &
Stage-1 Routing &
Stage-2 Scope &
Evidence Structure &
Packing under $B_{\text{tok}}$ \\
\midrule

\multicolumn{8}{l}{\textbf{Text chunk-based RAG}} \\
\addlinespace[0.2em]

Page (Text-only) &
Page &
Text (OCR) &
None &
N &
Corpus-wide &
None &
Fixed top-$K$ pages (page-budget) \\

Length chunking~\citep{Gong2020Recurrent} &
Chunk &
Text (OCR) &
None &
N &
Corpus-wide &
None &
Greedy top-$k$ chunks \\

LumberChunker~\citep{duarte-etal-2024-lumberchunker} &
Chunk &
Text (OCR) &
None &
N &
Corpus-wide &
None &
Greedy top-$k$ chunks \\

Meta Chunker~\citep{zhao-etal-2025-moc} &
Chunk &
Text (OCR) &
None &
N &
Corpus-wide &
None &
Greedy top-$k$ chunks \\

Structural chunking~\citep{yepes2024financialreportchunkingeffective} &
Chunk &
Text (OCR) &
Weak (layout-based boundaries) &
N &
Corpus-wide &
None &
Greedy top-$k$ chunks \\

MultiDocFusion~\citep{multidocfusion} &
Hier. chunk &
Text (OCR) &
Native / structure-aware &
N (no \texttt{Hierarchy Field}) &
Corpus-wide &
No explicit graph (chunk hierarchy only) &
Greedy top-$k$ chunks (no subgraph assembly) \\

\midrule
\multicolumn{8}{l}{\textbf{Text-based GraphRAG}} \\
\addlinespace[0.2em]

RAPTOR~\citep{sarthi2024raptorrecursiveabstractiveprocessing} &
Chunk + Summary &
Text &
\textit{Induced} (summary tree) &
N &
Corpus-wide &
Recursive summary tree &
Greedy top-$k$ nodes \\

HopRAG~\citep{liu-etal-2025-hoprag} &
Chunk &
Text &
None &
N &
Corpus-wide &
Chunk graph (similarity / hop expansion) &
Greedy top-$k$ chunks \\

\midrule
\multicolumn{8}{l}{\textbf{Page-level multimodal RAG}} \\
\addlinespace[0.2em]

M3DocRAG~\citep{Cho_2025_ICCV} &
Page &
Page-level MM &
None &
N &
Corpus-wide &
None &
Fixed top-$K$ pages (page-budget) \\

VDocRAG~\citep{vdocrag2025} &
Page &
Page-level MM &
None &
N &
Corpus-wide &
None &
Fixed top-$K$ pages (page-budget) \\

\midrule
\multicolumn{8}{l}{\textbf{Multimodal GraphRAG}} \\
\addlinespace[0.2em]

MoLoRAG~\citep{wu-etal-2025-molorag} &
Page &
Page-level MM &
None &
N &
Corpus-wide &
Page graph expansion &
Fixed top-$K$ pages (page-budget) \\

SimpleDoc~\citep{jain-etal-2025-simpledoc} &
Page &
Page-level MM &
None &
N &
Corpus-wide &
Page-level graph &
Fixed top-$K$ pages (page-budget) \\

\midrule
\ours{} &
Fine-grained MM unit &
Text + Visual Crop &
Native (DHP path) &
Y: Hierarchy Routing &
Cand. docs + Anchor subtree &
Heterogeneous Tree Graph (tree edges only) &
Ancestry-aware hybrid packing (Sibling + Semantic) \\

\bottomrule
\end{tabular}
\end{adjustbox}

\caption{
Qualitative comparison of retrieval framework design choices. We contrast methods by retrieval unit, retriever/modality, hierarchy usage and Stage-1 routing, Stage-2 search scope, evidence structure (none vs. graph), and packing under a token budget $B_{\text{tok}}$.
}
\label{tab:axis_comparison_hikey_vs_baselines}
\end{table*}

For fair comparison, all baselines use the same source corpus, corpus split, Reader, and decoding settings. Each method constructs its own retrieval index according to its retrieval unit and modality assumptions.
Differences arise primarily from retrieval unit choice, routing strategy, and modality utilization.

\subsection{Text Chunk-based RAG}
The most common RAG setup: flat retrieval over text units without hierarchical routing.

\paragraph{Page (Text-only).} Treats an entire page's extracted text as a single retrieval unit.
\paragraph{Length Chunking~\citep{Gong2020Recurrent}.} Mechanically splits text by fixed length (tokens/chars), often disrupting context.
\paragraph{LumberChunker~\citep{duarte-etal-2024-lumberchunker}.} Uses an LLM to detect topic shifts and set chunk boundaries dynamically based on semantic coherence.
\paragraph{Meta Chunker~\citep{zhao-etal-2025-moc}.} Merges chunks at the sentence/paragraph level by analyzing perplexity distributions.
\paragraph{Structural chunking~\citep{yepes2024financialreportchunkingeffective}.} Chunks based on layout types (headers, paragraphs), but unlike \ours{}, it does not reconstruct a full section tree or treat tables/figures as first-class retrieval nodes with ancestry-aware packing.
\paragraph{MultiDocFusion~\citep{multidocfusion}.} A strong hierarchy-aware baseline that preserves some structure, but lacks field-separated routing and multimodal unit fusion.

\subsection{Text-based GraphRAG}
Connects text units via graphs to enable multi-step exploration, but relies solely on text.

\paragraph{RAPTOR~\citep{sarthi2024raptorrecursiveabstractiveprocessing}.} Clusters and summarizes chunks to form a tree. Note that this \textit{induced} tree is semantic and not tied to the document's native layout structure.
\paragraph{HopRAG~\citep{liu-etal-2025-hoprag}.} Expands search along a similarity graph of text units using LLM-generated pseudo-queries, but does not natively handle tables or figures.

\subsection{Page-level Multimodal RAG}
Retrieves by embedding full-page images. While capturing layout, they suffer from coarse granularity.

\paragraph{M3DocRAG \& VDocRAG.}
M3DocRAG~\citep{Cho_2025_ICCV} and VDocRAG~\citep{vdocrag2025} both utilize multimodal embeddings but differ in retrieval formulation.
M3DocRAG employs a ColPali-style late-interaction retriever (token-level page embeddings scored by MaxSim). We adapt this to the ODQA setting by flattening all pages across the corpus into a single joint index.
In contrast, VDocRAG uses an LVLM-based dual-encoder that compresses each page image into a single dense representation (e.g., via EOS-token pooling) for maximum inner-product search.

\subsection{Multimodal GraphRAG}
Constructs graphs where nodes are full pages, limiting fine-grained control.

\paragraph{MoLoRAG~\citep{wu-etal-2025-molorag} \& SimpleDoc~\citep{jain-etal-2025-simpledoc}.}
These methods expand retrieval via page connectivity (e.g., similarity edges). However, the page-level granularity limits their ability to assemble a precise evidence subgraph under tight token budgets.

\section{Hierarchical Parsing, Section Path, and \texttt{Doc\_card}}
\label{sec:hikey_offline_details}

This section details the offline construction process of \ours{}: hierarchical parsing (DHP), section-path extraction, and the practical construction of \texttt{Doc\_card}.

\subsection{Document Hierarchical Parsing (DHP)}
\label{sec:dhp_details}

\begin{table*}[t]
\centering
\scriptsize
\renewcommand{\arraystretch}{1.12}
\setlength{\tabcolsep}{3.6pt}

\begin{adjustbox}{max width=\textwidth}
\begin{tabular}{lcccc}
\toprule
\textbf{Method} &
\textbf{HRDoc-S} &
\textbf{HRDoc-H} &
\textbf{DocHieNet} &
\textbf{Avg} \\
& \textbf{F1/STEDS} & \textbf{F1/STEDS} & \textbf{F1/STEDS} & \textbf{F1/STEDS} \\
\midrule

\multicolumn{5}{l}{\emph{Generic LVLM baselines (zero-shot image understanding)}}\\[-0.25em]
LLaVA-OneVision1.5~\citep{an2025llavaonevision15fullyopenframework}    & 27.61/12.93 & 26.30/18.21 & 17.78/\phantom{0}8.57 & 23.90/13.24 \\
InternVL3.5~\citep{wang2025internvl35advancingopensourcemultimodal}           & 28.40/14.47 & 27.57/19.98 & 18.18/\phantom{0}9.60 & 24.72/14.68 \\
Qwen2.5-VL~\citep{qwen25vl}            & 28.41/14.51 & 27.57/19.99 & 18.20/\phantom{0}9.62 & 24.73/14.71 \\

\midrule
\multicolumn{5}{l}{\emph{Structured hierarchy parsers (trained for tree reconstruction)}}\\[-0.25em]
DocParser~\citep{rausch2021docparser}             & 47.09/31.03 & 35.41/27.15 & 10.68/\phantom{0}4.31 & 31.06/20.83 \\
DSG~\citep{rausch2023dsgendtoenddocumentstructure}                   & 48.43/32.13 & 36.42/27.69 & 26.71/19.45 & 37.19/26.42 \\
DSPS~\citep{10.1609/aaai.v37i2.25277}                  & 65.27/59.57 & 54.06/38.41 & 35.61/23.81 & 51.65/40.60 \\
DSHP-LLM~\citep{multidocfusion}              & 44.90/29.52 & 61.29/51.34 & 64.29/53.49 & 56.83/44.78 \\
Qwen2.5-VL--DHP--SFT  & 50.97/46.75 & 43.05/41.02 & 42.85/40.39 & 45.62/42.72 \\

\midrule
\textbf{M3DocDep (DHP used in HiKEY)} &
\textbf{82.87/76.52} &
\textbf{77.75/71.65} &
\textbf{76.01/70.83} &
\textbf{78.88/72.99} \\
\bottomrule
\end{tabular}
\end{adjustbox}

\caption{
Standalone validation of the DHP backbone used in HiKEY, adapted from M3DocDep~\citep{shin2026m3docdep}. We use the same DHP checkpoint as the upstream hierarchy parser in HiKEY, and report F1 and STEDS on HRDoc-S, HRDoc-H, and DocHieNet.
}

\label{tab:dhp_backbone_validation_appendix}
\end{table*}

\paragraph{Role in HiKEY.}
HiKEY relies on an upstream Document Hierarchical Parsing (DHP) module to recover document structure from raw PDFs.
Given a document $d$, DHP produces (i) a hierarchy tree $\mathcal{T}(d)$ over layout blocks (e.g., headings, paragraphs, captions, tables, and figures), which provides reliable section paths for Stage-1 routing and ancestry context for evidence packing.
Concretely, the recovered section path is serialized into the hierarchy field of each Doc\_card/Sec\_card, enabling query routing by document-level structure rather than flat chunk similarity.

\paragraph{LVLM-based multimodal DHP.}
A key distinction from prior DHP-style components is that our DHP is an \emph{LVLM-based multimodal} parser: it jointly consumes text, table crops, and figure crops as first-class inputs to hierarchy recovery.
By contrast, the closest prior components are typically LLM-based text-only parsers that cannot leverage visual signals (e.g., a table whose structure is visible only from its rendered layout, or a figure whose caption is laid out across columns).
This design choice is load-bearing for \ours{}: because downstream retrieval units include tables and figures as first-class nodes, the hierarchy over those units has to be recovered from a multimodal, not text-only, signal.

\paragraph{Model and outputs.}
We implement DHP with a structure-aware parser that operates on page-level layout blocks and predicts parent--child relations to reconstruct $\mathcal{T}(d)$, while also identifying block types and span boundaries required by downstream indexing.
All retrieval experiments in this paper use the same fixed DHP checkpoint; DHP is not tuned on ODQA benchmarks, ensuring that downstream gains are attributable to the retrieval framework rather than task-specific retraining of the parser.

\paragraph{Standalone validation of DHP reliability.}
A key concern is whether hierarchy recovery is accurate enough to support routing and structured packing.
We therefore include component-level validation results for the same DHP checkpoint used in HiKEY, adapted from M3DocDep~\citep{shin2026m3docdep}.
Appendix Table~\ref{tab:dhp_backbone_validation_appendix} reports F1 and STEDS~\citep{zhong2020image} on HRDoc-S, HRDoc-H~\citep{10.1609/aaai.v37i2.25277}, and DocHieNet~\citep{xing2024dochienet}.
These results are included only to characterize the reliability of the upstream parser; all ODQA retrieval and QA results in this paper are evaluated within the HiKEY pipeline.

\paragraph{Downstream sensitivity.}
Beyond standalone parsing scores, we additionally quantify the importance of hierarchy signals for retrieval.
Our ablation study (Table~6) shows that removing hierarchy-aware indexing/routing substantially degrades retrieval quality compared to field-separated hierarchy indexing with coarse-to-fine routing.
This indicates that HiKEY's improvements are not merely due to larger contexts or stronger encoders, but critically depend on accurate structural signals supplied by DHP.

\paragraph{Representative DHP failure cases and downstream impact.}
\label{par:dhp_failure_cases}
We catalog two failure modes that dominated the residual errors on HRDoc-S/H and DocHieNet and we trace their effect on HiKEY's retrieval and QA.
(\textbf{F-H}) \emph{Missing or misclassified headers.}
When a header block is detected as a paragraph (or vice versa), the affected unit's Governing Header defaults to a higher ancestor (or the document Title), which widens the section path.
Downstream, this hurts Stage-1 routing in two ways: the \texttt{Doc\_card} hierarchy field loses a discriminative section string, and the affected Sec\_card is merged with its neighbors, reducing Stage-2 ranking precision.
In our error inspection, documents with frequent header drops tended to show lower Recall@1 than documents with comparable body-text retrieval quality.
(\textbf{F-N}) \emph{Incorrect nesting (wrong parent).}
When a deeper header is attached to the wrong ancestor (e.g., a subsection nested under a sibling section), the resulting section path is structurally plausible but semantically mis-scoped.
This does not noticeably affect retrieval (the routing signal is coarse), but it harms Reader interpretability: during Phase~1 packing, the Anchor Unit is shipped with a path that points to the wrong section, which can mislead the Reader on queries that depend on scope (``under which section does X fall?'').
Ancestry-aware packing mitigates this partially by always attaching the full path; the Reader at least sees the candidate ancestor chain.
Both failure modes are observable in our standalone parsing metrics (STEDS captures nesting errors; F1 captures header-type errors), which is why we retain both numbers in Table~\ref{tab:dhp_backbone_validation_appendix} rather than reporting a single aggregate.

\paragraph{DHP is an offline, one-time indexing cost.}
\label{par:dhp_offline}
We stress that DHP is executed once per document during offline indexing, not at query time.
Once $\mathcal{T}(d)$ and the section paths are computed, they are persisted in the index; every subsequent query reuses them without re-running DHP.
The end-to-end latency a deployed HiKEY system exposes at serving time is therefore bounded by Stage-1 routing + Stage-2 scoring + Reader inference, none of which invoke DHP.
The runtime breakdown in Appendix~\ref{sec:runtime_supp} (Table~\ref{tab:supp_runtime}) reports this offline cost explicitly, separated from query-time latency; the per-document DHP cost amortizes over the query stream in any realistic deployment.

\paragraph{Discussion.}
DHP is an upstream module and may be imperfect on severely degraded scans or atypical layouts; however, HiKEY is designed to be robust by combining hierarchy-based routing with multimodal fine retrieval and budgeted evidence packing.
The standalone validation in Table~\ref{tab:dhp_backbone_validation_appendix} bounds the parser's average accuracy, the downstream ablations in Table~\ref{tab:ablation_hikey_avg_r10_onecol} isolate the contribution of the hierarchy signal within the pipeline, and the offline cost argument above addresses deployment efficiency.
Together, these address concerns about the reliability and practical cost of hierarchy parsing in our retrieval pipeline.

\subsection{Section-path Extraction}
\label{sec:section_path_details}

\paragraph{Governing Header.}
For each evidence unit (block) $c$, we traverse upward in the recovered tree $\mathcal{T}(d)$ and select the nearest \texttt{Title} or \texttt{Section Header} ancestor as its \emph{governing header}.

\paragraph{Section Path.}
Using the header sequence from the document root to the governing header, we construct the section path:
{\small
\[
\texttt{section\_path}(c) = \texttt{Title} > \texttt{Sec} > \texttt{Subsec} > \dots
\]
}
This path is stored as structural metadata and is utilized as a key feature for both the hierarchy field index and graph traversal.

\subsection{\texttt{Doc\_card} Construction}
\label{sec:doc_card_details}

Stage-1 routing relies on a lightweight summary representation, \texttt{Doc\_card}(d), designed to provide global topic cues without the noise of the full body text.

\paragraph{Deterministic Construction Rules.}
To implement the theoretical definition in Sec.~\ref{sec:method} efficiently, we construct the \texttt{Doc\_card} by concatenating:
(i) The detected \texttt{Title} text (selected via DP confidence and position),
(ii) A reading-order list of high-level section headers (typically depth 1--2),
(iii) Optionally, ToC entries if a Table of Contents page is identifiable.
For indexing stability, the final string is truncated to a fixed maximum length (tuned on the validation set).

\paragraph{Design Intent.}
This construction keeps the Stage-1 index compact while supplying strong global topic signals, preventing the routing module from being overly sensitive to noisy body OCR or irrelevant local details.

\subsection{No Explicit Cross-Block Link Modeling in HiKEY}
\label{sec:no_ref_edges}

HiKEY does not model, predict, or construct any explicit cross-block link graph beyond the hierarchy tree recovered by DHP.
In our pipeline, DHP is used solely to reconstruct the parent--child hierarchy $\mathcal{T}(d)$ (tree edges) and to provide ancestry context (section paths and governing headers).
All cross-unit association needed for ODQA is handled deterministically at packing time via the hybrid policy described below.

\subsection{Semantic Associate Mining for Hybrid Packing}
\label{sec:semantic_associate_mining}

For each Stage-2 anchor unit $c_i$, we mine Semantic Associates by ranking other units within the same document using a similarity function $Sim(c_i, \cdot)$ computed from precomputed dense embeddings:
text embeddings for textual units, and visual embeddings for table/figure crops.
During packing, we add high-similarity visual units as Semantic Associates subject to the token budget, and attach the required ancestry context from the DHP tree to keep each added unit interpretable in its original section context.

\section{Ancestry-aware Evidence Subgraph Assembly}
\label{sec:packing_details}

This section details the ancestry-aware packing policy used to assemble the final multimodal context under a strict token budget $B_{\text{tok}}$.

\subsection{Serialization Format}
\label{sec:serialization_details}

To ensure the Reader model perceives the logical structure, each selected evidence unit is serialized with its structural context:
\begin{itemize}[leftmargin=*]
    \item Ancestry Context: The document Title and the sequence of governing headers (from the DHP tree) required to locate the unit (e.g., \texttt{\# 2. Methods > \#\# 2.1. Model}).
    \item Unit Metadata: Unique identifiers, unit type (Text/Table/Figure), and source page number.
    \item Content: The unit's textual content (OCR text, linearized table, or caption).
    \item Visual Crop: For \texttt{Table} and \texttt{Figure} units, the corresponding image crop is inserted (if the LVLM budget allows).
\end{itemize}

\subsection{Ancestry-aware Packing Algorithm}
\label{sec:packing_algo}

Unlike naive greedy packing, our strategy explicitly prioritizes the structural integrity of evidence. We include high-scoring Stage-2 units first, mandatorily attach their ancestry nodes, and then expand using (i) Sibling Units under the same parent section and (ii) Semantic Associates selected by embedding similarity, to form a coherent subgraph under the token budget.

\paragraph{What does the ``subgraph'' contain?}
Each assembled evidence subgraph $\mathcal{S}$ is a set of nodes drawn from the DHP tree $\mathcal{T}(d)$ together with their ancestry chains, grouped into three disjoint roles:
(1) one or more \emph{Anchor Units} (the highest-ranked Stage-2 units for the query);
(2) the \emph{Governing Headers} of each included unit (the chain of ancestral section titles from the document root);
(3) optional \emph{Sibling Units} under the same parent section as an Anchor (Phase~2) and \emph{Semantic Associates} retrieved from elsewhere in the document by embedding similarity to an Anchor (Phase~3).
The subgraph is \emph{not} a newly learned cross-block link structure: edges are exactly the parent--child tree edges of $\mathcal{T}(d)$ restricted to the included nodes, with no predicted cross-unit links (Appendix~\ref{sec:no_ref_edges}).

\begin{figure}[t]
\centering
\begin{tikzpicture}[
  font=\scriptsize,
  node distance=2mm and 2mm,
  phase/.style={draw,rounded corners=1.2pt,text width=0.88\linewidth,align=center,inner sep=3pt},
  phase1/.style={phase,fill=blue!7, draw=blue!55!black},
  phase2/.style={phase,fill=green!7,draw=green!45!black},
  phase3/.style={phase,fill=orange!8,draw=orange!65!black},
  arr/.style={-{Latex[length=1.6mm]},thick,gray!70!black},
]
  \node[phase1] (p1) {%
    \textbf{Phase 1: Anchor \& Ancestry}\\[-1pt]
    {\tiny Insert ranked anchor $c_i$ + Governing Headers $\texttt{path}(c_i)$.
    Token cost $\Delta_{\text{anc}}$; skip $c_i$ if $C+\Delta_{\text{anc}}>B_{\text{tok}}$.}
  };
  \node[phase2, below=2mm of p1] (p2) {%
    \textbf{Phase 2: Sibling Packing (Structural)}\\[-1pt]
    {\tiny For each sibling $s$ under the same parent section as $c_i$: add with the inherited Ancestry if $C+\Delta_{\text{sib}}\le B_{\text{tok}}$.
    Preserves visual--text alignment within the subtree.}
  };
  \node[phase3, below=2mm of p2] (p3) {%
    \textbf{Phase 3: Semantic Packing (Cross-section)}\\[-1pt]
    {\tiny Top-$M$ units by $\mathrm{Sim}(c_i,\cdot)$ across the document; add visual units with their Ancestry if budget allows.
    Captures latent cross-section links.}
  };

  \draw[arr] (p1.south) -- (p2.north);
  \draw[arr] (p2.south) -- (p3.north);
\end{tikzpicture}
\caption{Schematic of the three packing phases in Algorithm~\ref{alg:packing_ancestry}. Each phase is budget-gated; the loop iterates over ranked Stage-2 anchors.}
\label{fig:packing_schematic}
\end{figure}
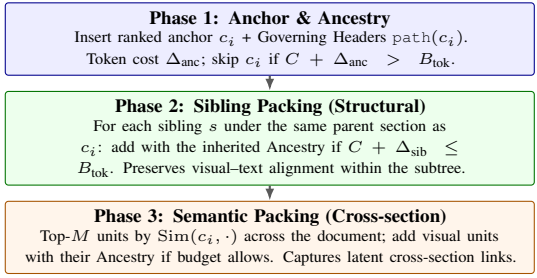

\paragraph{Schematic of the packing phases.}
Figure~\ref{fig:packing_schematic} visualizes the three phases of Algorithm~\ref{alg:packing_ancestry} for a single Stage-2 anchor.
Phase~1 seats the Anchor Unit together with its Governing Headers.
Phase~2 draws Sibling Units from the same parent subtree (structural association).
Phase~3 draws Semantic Associates from elsewhere in the document via embedding similarity (semantic association).
Each phase is gated by the remaining token budget $B_{\text{tok}}-C$; any unit that would overflow is skipped and the loop advances to the next ranked anchor.

\paragraph{Compact running example.}
We instantiate the three phases on the Apollo~11 query from Appendix~\ref{sec:qualitative_analysis} (\emph{``Who were the crew members of Apollo~11, and when did they land on the Moon?''}).
Assume the top-ranked Stage-2 anchor is the crew paragraph $c_1$ (Text) under section \texttt{3.1 Prime crew}.
\begin{itemize}[leftmargin=1.2em,itemsep=1pt,topsep=2pt]
  \item \emph{Phase 1 (Anchor + Ancestry)} seats $c_1$ together with its Governing Headers \texttt{Apollo~11 > 3 Mission personnel > 3.1 Prime crew} ($\Delta_{\text{anc}}\!\approx\!180$ tokens under the Qwen2.5-VL tokenizer).
  \item \emph{Phase 2 (Sibling expansion)} adds the crew table $c_2$ and the crew portrait figure $c_3$; both are Sibling Units under \texttt{3.1 Prime crew} (each $\Delta_{\text{sib}}\!\approx\!350$--$450$ tokens including vision-encoder patches).
  \item \emph{Phase 3 (Semantic expansion)} detects that the landing-time table under \texttt{5.1 Landing} has high visual--semantic similarity to $c_1$ (the query also asks about ``landing''); it is added as a Semantic Associate, and its own Governing Headers \texttt{5 Mission events > 5.1 Landing} are attached so the Reader can locate it.
\end{itemize}
The assembled subgraph is thus $\mathcal{S} = \{c_1, c_2, c_3, c_{\text{landing}}\}$ together with two ancestry chains, and is serialized into the Reader prompt as shown in Appendix~\ref{sec:qualitative_analysis}.
Critically, the Reader sees each unit together with its own ancestry, so a cross-section unit added via Phase~3 does not cause scope confusion -- it is explicitly tagged with the section path \texttt{5.1 Landing}, distinct from the anchor's path \texttt{3.1 Prime crew}.

\begin{algorithm}[t]
\footnotesize
\caption{An ancestry-aware hybrid packing algorithm that assembles a coherent evidence subgraph using Sibling and Semantic strategies.}
\label{alg:packing_ancestry}
\begin{algorithmic}[1]
\Require Stage-2 ranked anchors $\{(c_i, s_i)\}_{i=1}^N$, DHP tree $\mathcal{T}(d)$, semantic similarity function $Sim(\cdot, \cdot)$, token budget $B_{\text{tok}}$
\Ensure Evidence subgraph $\mathcal{S}$
\State $\mathcal{S} \gets \emptyset$, $C \gets 0$ \Comment{$C$: current accumulated token count}
\For{$i=1$ to $N$}
  \State Phase 1: Anchor \& Ancestry
  \State $\Delta_{\text{anc}} \gets$ tokens for unit $c_i$ + Ancestry Context
  \If{$C + \Delta_{\text{anc}} > B_{\text{tok}}$} continue \EndIf
  \State Add $c_i$ and Ancestry to $\mathcal{S}$; $C \gets C+\Delta_{\text{anc}}$

  \State Phase 2: Sibling Packing (Structural)
  \State $Siblings \gets$ \{units sharing same parent section with $c_i$\}
  \For{$s \in Siblings$}
    \State $\Delta_{\text{sib}} \gets$ tokens for $s$ \Comment{Sibling $s$ inherits the same Ancestry Context as $c_i$}
    \If{$C + \Delta_{\text{sib}} \le B_{\text{tok}}$ and $s \notin \mathcal{S}$}
       \State Add $s$ (with inherited Ancestry) to $\mathcal{S}$; $C \gets C+\Delta_{\text{sib}}$
    \EndIf
  \EndFor

  \State Phase 3: Semantic Packing (Cross-Section)
  \State $Candidates \gets$ Top-$M$ units by $Sim(c_i, \cdot)$ from Doc
  \For{$m \in Candidates$}
    \If{$m$ is visual unit and $m \notin \mathcal{S}$}
       \State $\Delta_{\text{sem}} \gets$ tokens for $m$ + Ancestry
       \If{$C + \Delta_{\text{sem}} \le B_{\text{tok}}$}
          \State Add $m$ and Ancestry to $\mathcal{S}$; $C \gets C+\Delta_{\text{sem}}$
       \EndIf
    \EndIf
  \EndFor
\EndFor
\State \Return $\mathcal{S}$
\end{algorithmic}
\end{algorithm}

\paragraph{Implementation Note.}
Token counting is performed using the specific tokenizer of the Reader model (Qwen2.5-VL).
Image crops are assigned a fixed token cost (corresponding to the vision encoder's patch tokens), and we strictly cap the total number of images to satisfy the LVLM's maximum input resolution constraints.

\section{LVLM and RAG Settings}
\label{sec:lvlm_rag_supp}

This section provides detailed experimental settings (retriever, reader, and hyperparameters) to ensure reproducibility. All experiments are implemented using PyTorch and HuggingFace Transformers.

\begin{table}[h]
\centering
\small
\renewcommand{\arraystretch}{1.25}
\setlength{\tabcolsep}{6pt}

\resizebox{\linewidth}{!}{%
\begin{tabular}{
  >{\raggedright\arraybackslash}p{0.35\linewidth}%
  >{\raggedright\arraybackslash}p{0.65\linewidth}}
\toprule
Setting & Configuration \\
\midrule
Infrastructure & NVIDIA H100 (80GB) $\times$ 4 \\
Index Unit & Corpus-level Joint Indexing \\
\midrule
Sparse Retriever & BM25 ($k_1=1.5, b=0.75$) \\
Dense Retriever (Text) & \texttt{gte-Qwen2-7b-instruct} \\
Dense Retriever (Visual) & \texttt{MM-Embed} \\
\midrule
Strategy & 
Stage-1: Hierarchical Routing (BM25 + Dense) \\
& Stage-2: Fine-grained Multimodal Fusion \\
\midrule
Reader Model & \texttt{Qwen2.5-VL-7B-Instruct} \\
Context Limit & 16,384 tokens \\
Decoding & Greedy decoding (Temperature=0.0) \\
\midrule
Evaluation Metrics & 
Retrieval: Recall@K, MRR@K, Hit@K, All@K \\
& QA: EM, ANLS, ROUGE-L, METEOR \\
\bottomrule
\end{tabular}
}
\caption{Experimental configuration summary for retrieval and QA. We list infrastructure, sparse/dense retrievers (text and visual), the two-stage strategy (hierarchical routing + fine-grained fusion), the LVLM reader configuration, and evaluation metrics.}
\label{tab:supp_rag_config}
\end{table}

\subsection{Retrieval Configuration}
\label{sec:retrieval_config_details}

\paragraph{Sparse Retrieval.}
We use standard BM25 settings from production search libraries (e.g., Pyserini/Elasticsearch).
After stopword removal, we use tokenized terms (e.g., from a morphological analyzer) for both Stage-1 routing fields and Stage-2 text units.

\paragraph{Dense Retrieval.}
We employ gte-Qwen2-7b-instruct~\citep{li2023towards} for text embeddings with an 8192-token window to handle long contexts.
For visual evidence units (tables/figures), we crop the regions and encode them using \texttt{MM-Embed}~\citep{lin2024nvmmembed} into 4096-dimensional embeddings.
Signals are combined via late fusion using learned weights tuned on the validation set.

\subsection{Reader Configuration}
\label{sec:reader_config_details}

\paragraph{Model \& Decoding.}
We use Qwen2.5-VL-7B-Instruct as the Reader.
For consistency and reproducibility, we fix Temperature to 0.0 (greedy decoding) and cap maximum generation at 256 tokens.

\paragraph{Input Context.}
The retrieved ancestry-aware evidence subgraph is serialized into a textual prompt structure. The total input length is strictly limited to 16K tokens to reflect realistic H100 memory constraints.

\subsection{Prompt Template}
\label{sec:prompt_templates}

The QA prompt template used in our experiments is illustrated in Fig.~\ref{fig:qa_prompt}. We strictly instruct the model to rely solely on the provided \texttt{[Context]} (the assembled subgraph) to minimize hallucinations.

\begin{figure}[h]
  \centering
  \begin{tcolorbox}[
    title={QA Prompt Template},
    colback=white,
    colframe=black,
    boxrule=0.8pt,
    arc=2pt,
    left=6pt,
    right=6pt,
    top=6pt,
    bottom=6pt,
    fonttitle=\bfseries
  ]
  \small
  [System Instruction] \\
  You are an AI assistant that answers questions by analyzing the provided documents. \\
  Write an accurate answer to \texttt{[Question]} using only the \texttt{[Context]} given below. \\
  - Do not answer using information that is not present in the context. \\
  - When referring to tables or figures, explicitly mention their IDs (e.g., Figure 3). \\
  - If you cannot be confident, output ``I do not have enough information to answer.''
   
  \vspace{0.3cm}
   
  [Question] \\
  \texttt{<QUESTION\_TEXT>}
   
  \vspace{0.3cm}
   
  [Context] \\
  \texttt{<SERIALIZED\_EVIDENCE\_SUBGRAPH>}
   
  \vspace{0.3cm}
  [Answer]
  \end{tcolorbox}
  \caption{The LVLM QA prompt template, where the [Context] slot is filled with the serialized ancestry-aware evidence subgraph assembled under the token budget.}
  \label{fig:qa_prompt}
\end{figure}

\definecolor{systemcolor}{HTML}{E8F1FF}
\definecolor{tagblue}{HTML}{3B82F6}
\definecolor{tagred}{HTML}{EF4444}
\definecolor{taggreen}{HTML}{10B981}
\definecolor{imageplaceholder}{HTML}{F3F4F6}
\definecolor{codebg}{HTML}{F8F9FA}

\section{Qualitative Case Study and Analysis}
\label{sec:qualitative_analysis}

This section analyzes how \ours{} processes Wikipedia-style multimodal documents. We trace how the pipeline in Fig.~\ref{fig:system_architecture} applies to an \emph{``Apollo 11''} document (step-by-step walkthrough) and visualize the final evidence subgraph serialized for the LLM Reader.

\subsection{System Walkthrough}
When a user asks: \emph{``Who were the crew members of Apollo 11, and when did they land on the Moon?''}, \ours{} generates an answer via the following internal steps.

\begin{tcolorbox}[
  enhanced,
  breakable,
  colback=systemcolor,
  colframe=black!30,
  title=Inference Walkthrough: Apollo 11 Mission Query,
  fonttitle=\bfseries\small,
  boxrule=0.6pt,
  arc=2mm,
  left=2mm, right=2mm, top=2mm, bottom=2mm,
  before skip=6pt,
  after skip=6pt
]
\footnotesize
(a) Document Parsing \& OCR.\par
We parse \texttt{Apollo\_11.pdf} and extract layout blocks.
\begin{itemize}[leftmargin=1.5em, nosep]
  \item \texttt{B\_h}: Heading ``3. Mission personnel'' / ``5.1 Landing''
  \item \texttt{B\_p}: Text ``The prime crew selected for Apollo 11 consisted of...''
  \item \texttt{B\_t}: Table ``Position | Astronaut'' (Commander: Neil Armstrong...)
  \item \texttt{B\_f}: Figure (Official Crew Portrait) (+ Crop)
\end{itemize}

\smallskip
(b) Hierarchy \& Graph Construction.\par
The DHP model reconstructs the ToC-like structure and assigns section paths.
\begin{itemize}[leftmargin=1.5em, nosep]
  \item Path: $\texttt{Apollo 11} \rightarrow \texttt{3 Mission personnel} \rightarrow \texttt{3.1 Prime crew}$
  \item Units: $c_1(\text{Text})$, $c_2(\text{Table})$, $c_3(\text{Figure})$
  \item Edges: Tree($\texttt{Sec 3.1} \rightarrow c_1, c_2, c_3$)
\end{itemize}

\smallskip
(c) Stage-1: Hierarchical Routing.\par
We match query keywords (``crew'', ``landing'') against the hierarchy field (\texttt{Doc\_card}) to narrow the scope.
\begin{itemize}[leftmargin=1.5em, nosep]
  \item Output: Target Doc=$\{\texttt{Apollo\_11.pdf}\}$, Anchors=$\{\texttt{Sec 3.1}, \texttt{Sec 5.1}\}$
\end{itemize}

\smallskip
(d) Stage-2: Fine Retrieval \& Assembly.\par
We retrieve units under the anchors and perform ancestry-aware packing of the text ($c_1$), the crew table ($c_2$), and the landing-time table, strictly adhering to the token budget $B_{\text{tok}}$. The crew table ($c_2$) is included as a Sibling Unit, and the landing-time table is retrieved as a Semantic Associate.
\end{tcolorbox}

\subsection{Serialized Input Visualization}
The selected evidence units are serialized and fed to the LVLM. The example below illustrates the coherent integration of textual explanation with the crew table and a linked crew photo.

\definecolor{tagblue}{HTML}{3B82F6}
\definecolor{tagred}{HTML}{EF4444}
\definecolor{taggreen}{HTML}{10B981}
\definecolor{imageplaceholder}{HTML}{F3F4F6}

\begin{tcolorbox}[
    enhanced,
    breakable,
    width=\columnwidth,
    colback=white,
    colframe=gray!60,
    title=Example of Serialized Evidence Subgraph (Input Prompt),
    fonttitle=\bfseries\small,
    boxrule=0.8pt,
    arc=2pt,
    left=3pt, right=3pt, top=3pt, bottom=3pt,
    before skip=6pt,
    after skip=8pt
]
\ttfamily\fontsize{7.3pt}{8.7pt}\selectfont

\textcolor{gray}{\# Global Context}\par
[DOC\_META]\par
ID: \textcolor{tagblue}{Apollo\_11.pdf} | Title: Apollo 11 - Wikipedia

\smallskip
\rule{\linewidth}{0.4pt}
\smallskip

\textcolor{gray}{\# Hierarchy Anchor (Routing Result)}\par
\textcolor{tagblue}{[SCOPE]} \textcolor{black}{3. Mission personnel > 3.1. Prime crew}

\medskip

\textcolor{taggreen}{[UNIT id=42 | Type=Text]}\par
Path: 3. Mission personnel > 3.1. Prime crew\par
Content: The prime crew for Apollo 11 consisted of Commander Neil Armstrong, Command Module Pilot Michael Collins, and Lunar Module Pilot Edwin "Buzz" Aldrin. Their positions are detailed in \textcolor{tagred}{Table 1}.

\medskip

\textcolor{taggreen}{[UNIT id=43 | Type=Table]}\par
Path: 3. Mission personnel > 3.1. Prime crew\par
Visual: \texttt{<|image\_token\_1|>} (Table Crop)\par
\begin{tikzpicture}
    \node[
        draw=gray!30,
        fill=imageplaceholder,
        inner sep=0pt,
        minimum width=0.98\linewidth,
        minimum height=1.2cm,
        align=center,
        text=gray!70,
        font=\sffamily\footnotesize\itshape,
        rounded corners=1pt
    ] {
        [ IMAGE PLACEHOLDER ] \\
        (Cropped image of Table 1: Prime Crew Members)
    };
\end{tikzpicture}

\medskip

\hfill \textit{\textcolor{tagblue}{$\Downarrow$ Sibling Evidence (Same Section) $\Downarrow$}} \hfill\mbox{}

\medskip

\textcolor{taggreen}{[UNIT id=45 | Type=Figure]}\par
Path: 3. Mission personnel > 3.1. Prime crew\par
Caption: Figure 2. The official crew portrait of Apollo 11.\par
Visual: \texttt{<|image\_token\_2|>} (Photo Crop)\par
\begin{tikzpicture}
    \node[
        draw=gray!30,
        fill=imageplaceholder,
        inner sep=0pt,
        minimum width=0.98\linewidth,
        minimum height=1.5cm,
        align=center,
        text=gray!70,
        font=\sffamily\footnotesize\itshape,
        rounded corners=1pt
    ] {
        [ IMAGE PLACEHOLDER ] \\
        (Cropped photo of Armstrong, Collins, and Aldrin)
    };
\end{tikzpicture}

\end{tcolorbox}

\section{Qualitative Failure Cases of Baseline Families}
\label{sec:failure_cases}

We summarize three representative failure modes repeatedly observed when inspecting baseline outputs on FRAMES and M3DocVQA.
These are intended to make concrete the structural limits captured by Table~\ref{tab:why_baselines_fail} in the main text; we do not claim they are exhaustive.

\paragraph{(F1) Chunk-RAG: routing locked onto a lexically similar but topically wrong document.}
For composite FRAMES queries that share named entities across unrelated articles (e.g., several Wikipedia pages that mention ``Apollo''), flat chunk retrievers rank chunks from a nearby but incorrect document at the top because the similarity signal is dominated by local token overlap.
Once the wrong document is selected, subsequent chunks reinforce the mistake, which is one reason chunk-RAG's FRAMES \emph{All} score is substantially below its Hit score.
\ours{} avoids this by routing at the \texttt{Doc\_card} level, where the hierarchy field (section paths) separates topically different documents even when their body text looks similar.

\paragraph{(F2) Page-level multimodal RAG: correct page retrieved, answer drowned by page-wide noise.}
Page-embedding methods often retrieve the correct page but then fail in end-to-end QA because the fixed page unit injects large amounts of unrelated text and visual noise into the 16K budget, pushing the actual answer span below the model's attention priority.
This is consistent with the gap between their retrieval Hit@K and end-to-end EM in Tables~\ref{tab:doc_retrieval_avg_1_10} and \ref{tab:main_e2e_qa}.
\ours{} instead packs a block-level Anchor Unit with only the ancestry needed for interpretation, preserving budget for sibling/semantic expansion.

\paragraph{(F3) Text-only GraphRAG: multi-hop succeeds on text but breaks on table/figure-anchored evidence.}
RAPTOR and HopRAG connect text chunks but do not expose tables and figures as first-class retrieval targets; for M3DocVQA questions whose answer lives in a table row or a figure caption, these systems at best retrieve a nearby text chunk that \emph{mentions} the table, missing the content itself.
\ours{}'s Sec\_card treats the table/figure crop as the retrieval unit with Upper Context attached, so the answer-bearing unit is ranked and packed directly (see Tab.~\ref{tab:mmqa_method_breakdown_source_hop} for the Table/Image breakdown).

\section{Runtime and Scalability Analysis}
\label{sec:runtime_supp}

We analyze the computational cost of \ours{} under a realistic industrial setting.
Measurements are conducted on a single NVIDIA H100 (80GB) GPU with 144 DPI rendering and high-precision OCR settings; peak GPU memory during offline indexing is approximately 27~GB.

\subsection{Offline Indexing Cost}
Table~\ref{tab:supp_runtime} breaks down the end-to-end processing runtime by document length, including high-resolution layout detection, OCR, visual embedding, DHP tree parsing, and graph construction.
The full offline pipeline amortizes to roughly 8--9~s/page on H100 in this high-precision configuration; the core DHP tree parsing itself is lightweight (about 0.1~s/page), so the bulk of the cost is paid once by OCR and visual embedding at index time, not at query time.

\begin{table}[!t]
\centering
\scriptsize
\renewcommand{\arraystretch}{1.2}
\setlength{\tabcolsep}{2.0pt}

\resizebox{\columnwidth}{!}{%
\begin{tabular}{@{}l ccc c@{}}
\toprule
\multirow{2}{*}{Stage} & \multicolumn{3}{c}{Latency (s)} & \multirow{2}{*}{Complexity} \\
\cmidrule(lr){2-4}
& 5 pages & 10 pages & 20 pages & \\
\midrule
\multicolumn{5}{l}{\textit{Visual Analysis (High-Res)}} \\
\quad Layout Detection & 5.2 & 10.5 & 21.2 & $O(P)$ \\
\quad OCR (Dense Text) & 24.5 & 48.2 & 98.4 & $O(P \times T)$ \\
\quad Visual Embedding & 12.5 & 25.8 & 52.6 & $O(N_{\text{img}})$ \\
\midrule
\multicolumn{5}{l}{\textit{Structure \& Graph Construction}} \\
\quad DHP Tree Parse & 0.5 & 1.0 & 2.1 & $O(N_{\text{blk}})$ \\
\quad Graph Build & 0.2 & 0.4 & 0.9 & $O(N + E)$ \\
\midrule
Total Indexing Time & 42.8 & 85.8 & 175.2 & Linear w.r.t Pages \\
\bottomrule
\end{tabular}%
}
\caption{
Offline indexing runtime and scalability as a function of document length. We report stage-wise latency (layout detection, OCR, visual embeddings, DHP parsing, and graph building) for 5/10/20-page documents and provide complexity terms indicating the main cost drivers.
}
\label{tab:supp_runtime}
\end{table}

\paragraph{Findings and Cost Justification.}
Over 90\% of the total indexing time is allocated to visual analysis (layout, OCR, embeddings). This reflects the unavoidable cost of preserving complex industrial document structures in high resolution.
Crucially, the core logic of \ours{}---tree parsing and graph construction---is extremely lightweight ($<2\%$ of total time) and does not create a bottleneck. This confirms that \ours{} efficiently organizes expensive vision-derived signals into a structured index without adding significant overhead.

\end{document}